\documentclass[sigconf, anonymous=False, balance=false]{acmart}

\usepackage{algorithmic}
\usepackage{algorithm}
\usepackage{amsfonts}
\usepackage{dsfont}
\usepackage{fancybox}
\usepackage{float}
    \usepackage{fvextra}
    \usepackage{gensymb} 
\usepackage{hyperref} 
\usepackage{listings}
\usepackage{multirow}
\usepackage{siunitx} 
\usepackage{subcaption}
\usepackage{textcomp}
    \usepackage{tikz}
    \usetikzlibrary {arrows.meta}
\usepackage{verbatim}
\usepackage{xcolor}

\AtBeginDocument{%
  }

\newcommand*\circled[1]{\tikz[baseline=(char.base)]{
            \node[shape=circle,draw,inner sep=1pt] (char) {#1};}}

\newcommand*\rectangled[1]{\tikz[baseline=(char.base)]{
            \node[shape=rectangle,draw,inner sep=1pt] (char) {#1};}}

\setcopyright{acmlicensed}
\copyrightyear{2018}
\acmYear{2018}
\acmDOI{XXXXXXX.XXXXXXX}

\acmConference[ACM e-Energy]{ACM International Conference on Future and Sustainable Energy Systems}{June 17 - 20, 2025}{Rotterdam, Netherlands}

\acmSubmissionID{17}

\begin{document}

\title{Large Language Model Interface for Home Energy Management Systems}

\author{François Michelon}
\orcid{0009-0002-1392-1100}
\affiliation{
    \institution{Ecole Centrale de Nantes}
    \city{Nantes}
    \country{France}}
\email{michelonfrancois00@gmail.com}

\author{Yihong Zhou}
\orcid{0000-0002-5015-8661}
\affiliation{
    \institution{University of Edinburgh}
    \department{School of Engineering}
    \city{Edinburgh}
    \country{United Kingdom}}
\email{yihong.zhou@eng.ox.ac.uk}

\author{Thomas Morstyn}
\orcid{0000-0003-2781-9588}
\affiliation{
    \institution{Oxford University}
    \department{Department of Engineering Science}
    \city{Oxford}
    \country{United Kingdom}}
\email{thomas.morstyn@eng.ox.ac.uk}

\begin{abstract}
    Home Energy Management Systems (HEMSs) help households tailor their electricity usage based on power system signals such as energy prices. This technology helps to reduce energy bills and offers greater demand-side flexibility that supports the power system stability. However, residents who lack a technical background may find it difficult to use HEMSs effectively, because HEMSs require well-formatted parameterization that reflects the characteristics of the energy resources, houses, and users' needs. Recently, Large-Language Models (LLMs) have demonstrated an outstanding ability in language understanding. Motivated by this, we propose an LLM-based interface that interacts with users to understand and parameterize their ``badly-formatted answers'', and then outputs well-formatted parameters to implement an HEMS. We further use Reason and Act method (ReAct) and few-shot prompting to enhance the LLM performance. Evaluating the interface performance requires multiple user--LLM interactions. To avoid the efforts in finding volunteer users and reduce the evaluation time, we additionally propose a method that uses another LLM to simulate users with varying expertise, ranging from knowledgeable to non-technical. By comprehensive evaluation, the proposed LLM-based HEMS interface achieves an average parameter retrieval accuracy of 88\%, outperforming benchmark models without ReAct and/or few-shot prompting.
\end{abstract}



\maketitle

\section{Introduction}
In the context of global decarbonization, the proportion of renewable energy sources such as wind and solar is increasing \cite{europa}. However, these renewable sources have intermittent generation, which challenges power system stability. Demand-side flexibility is acknowledged as an effective solution and is currently being implemented in practice.

Home Energy Management Systems (HEMS) represent a promising approach to enable demand-side flexibility \cite{STAVRAKAS2020112339}, which provides smart energy management for electricity generation, storage, and consumption in smart houses while respecting users' preferences \cite{5973168} \cite{ZHOU201630}. There have been extensive studies on HEMS. In \cite{MORSTYN2020115397} proposed an open-source software platform for integrated modeling, control, and simulation of smart local energy systems, including HEMS. In \cite{en9070562} proposed a method that includes users as key members of the HEMS, allowing them to plan the intended power consumption and manage real-time deviations. In \cite{Matsuyama2014} presented a grid-integrated system for energy management for smart demand-side energy management. The effectiveness of existing HEMS models was extensively discussed in \cite{LEE2016760}. In the literature, there has been extensive study of algorithm design for HEMS. However, the practical implementation of HEMS requires input parameters that represent the physical characteristics of home appliances, description of human behavior, and the user's needs for comfort. The required number of parameters can be high, and these must be well-formatted to be understandable to the HEMS algorithms. This process can be time-consuming and demotivates the use of HEMS \cite{CHEN2020101616}. Furthermore, non-expert users, particularly the elderly and those with limited literacy, may struggle to identify the correct and well-formatted parameters.

A new opportunity is presented by Large Language Models (LLMs), which perform particularly well in Natural Language Processing (NLP) tasks. This motivates us to use LLMs to retrieve well-formatted parameters required for HEMS, by interacting with users in an easy-understanding and natural-language way. Recently, LLMs have found applications for several power system tasks, including modeling nonlinear power electronic circuits \cite{zeid2024predictingdclinkcapacitorcurrent}, detecting insulator defect \cite{MAJUMDER20241544}, scheduling charging of electric vehicles (EVs) \cite{huang2023largefoundationmodelspower}, and performing power system simulation based on a user requests \cite{10251972, jia2024enablinglargelanguagemodels}. Using LLMs to facilitate the practical implementation of HEMS is a daily topic. Ref. \cite{designs8030043} presented an LLM-based system performing a discrete set of actions to manipulate home devices based on user instructions. Ref. \cite{10729865} proposed a smart home assistant that creates and controls home automation routines based on data sources. \cite{designs8030043} and \cite{10729865} implement systems that control home devices based on user instructions. However, the proposed systems require a lot of user interactions for simple tasks that will not permit efficient energy management. In this study, we propose an intuitive and flexible LLM-based interface that helps retrieve the necessary and well-formatted parameters for a HEMS to minimize household energy bills while fulfilling numerous user requirements. Specifically, we make the following contributions:
\begin{enumerate}
    \item We propose and develop a user-friendly LLM-based HEMS interface designed to fetch information through user--LLM interaction and convert it into a well-formatted form for optimization. This interaction, conducted in natural language, minimizes the effort and complexity of using HEMS.
    \item We propose the use of ReAct and few-shot prompting to improve the LLM performance, and this improvement is verified by case studies.
    \item Assessing the interface performance involves conducting multiple user--LLM tests. To minimize the effort in recruiting volunteers and decrease the evaluation time, we propose a method that employs an additional LLM to emulate users with diverse levels of expertise, from highly knowledgeable to non-technical.
\end{enumerate}

The manuscript is organized as follows. Section \ref{HEMS} describes HEMS and lists the main necessary inputs required for HEMS. Section \ref{methodology} introduces the proposed LLM-based interface and depicts its key features such as ReAct and few-shot prompting. 
Section \ref{user} describes LLM users by defining the cosine similarity metrics and the three user difficulty levels that will test the interface. Section \ref{results} presents the case study of the interface and the obtained results, showing that HEMS can be widespread when coupled with an intuitive and flexible interface. Section \ref{conclusion} concludes and discusses directions for future research.

\section{Home Energy Management System} \label{HEMS}
We consider a HEMS that minimizes the energy bill, by controlling devices such as electric heating, electric vehicle charging, user schedule, energy provider, solar panels, and simulation period and place (from which we obtain weather forecast and electricity price time series). This simple modelling can be extended to consider more detailed aspects like a more complex user schedule in future work. 

\subsection{Model definition}
The HEMS problem we consider optimizes the energy usage of flexible appliances over a time horizon to minimize the user's cost of energy, given constraints on how the appliances can be used. In our HEMS model, the user can possess several electric vehicles (EVs), the user has a daily schedule that defines when the user is home or not ($t_{arrival}$ when the user comes back to his accommodation and $t_{departure}$ when the user leaves). We assume that while parked at home the EV can be charged. When the user comes back from work, his EV has a low battery, and when the user leaves in the morning it is assumed his EV must be fully charged. In addition, the user also wants the house temperature to be between a minimum and maximum value. The model is more precisely described in the section \ref{hemsappendix} of the appendix.

\section{Methodology} \label{methodology}
\begin{figure}[ht!]
    \begin{center}
    \includegraphics[width=0.45\textwidth,height=0.22\textheight]{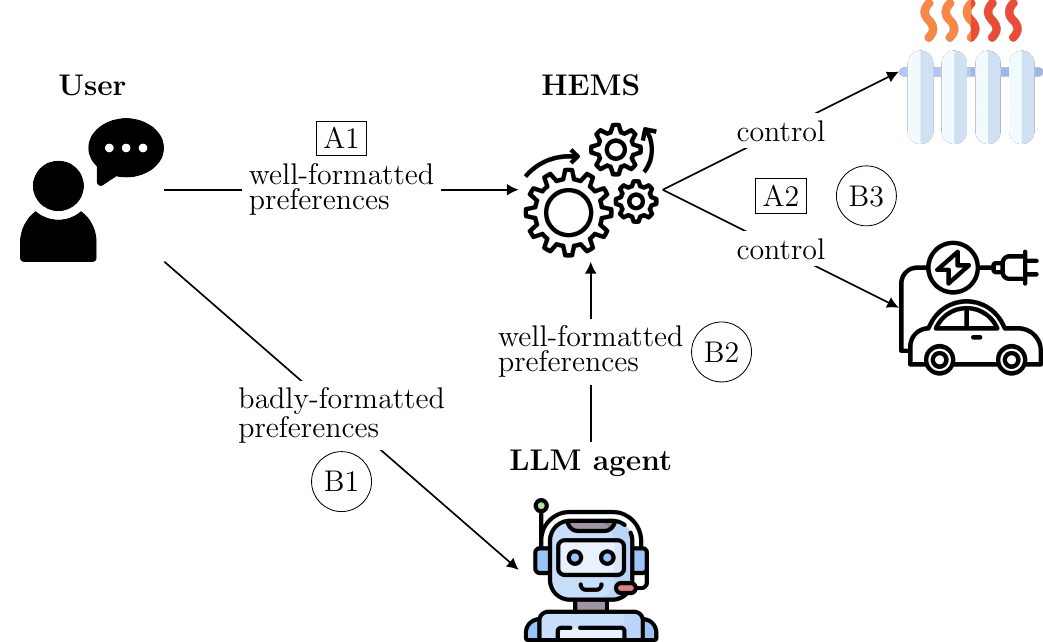}
    \caption[centerfirst]{High-level interface LLM-HEMS flowchart \par \small \normalfont Sub-flowchart \rectangled{A}, current Home Energy Management System (\textbf{HEMS}) requires well-formatted parameters from the \textbf{User}. After that, the \textbf{HEMS} controls the different appliances of the user's house to optimize the energy schedule while respecting users' preferences. Sub-flowchart \circled{B} illustrates the proposed LLM-integrated system. The \textbf{LLM agent} first interacts with the \textbf{user} by asking well-oriented questions. The \textbf{LLM agent} extracts the badly formatted preferences from the user's answers and converts them to fit the \textbf{HEMS} requirements. Finally, the \textbf{HEMS} can control the user's appliances as the user wants.}
    \label{fig:highlevelchart}
    \Description{Fully described in the caption.}
    \end{center}
\end{figure}

In most systems, users need to parameterize the HEMS by themselves (Fig. \ref{fig:highlevelchart} step \rectangled{\footnotesize{A1}}), which takes lots of time and can lead to input errors and control errors (Fig. \ref{fig:highlevelchart} step \rectangled{\footnotesize{A2}}). On the one hand, asking the user to provide too many precise parameters like EV capacity and usage schedule, will diminish the HEMS democratization as the learning curve will be too steep. On the other hand, asking for too little information will result in a non-tailored consumption energy schedule.

LLMs have exhibited versatile language understanding and prominent information retrieval ability in various contexts \cite{huang2023largefoundationmodelspower} \cite{jia2024enablinglargelanguagemodels}. Therefore, we propose to include LLM as an extra interface between users and HEMS. This paper considers a system with three main entities: the Home Energy Management System (HEMS), the LLM agent, and the user. We aim to design an LLM-based interface that retrieves all necessary user information, stores it in a correct format, and then sends the formatted inputs to the HEMS, to obtain the optimal electricity energy schedule. Fig. \ref{fig:highlevelchart} shows the high-level flowchart of our interface from user--LLM interaction where the user can provide badly-formatted parameters to the LLM like a textual description of its schedule instead of a formatted date (step \circled{\footnotesize{B1}}) to the parametrization of the HEMS with the translated user preferences into HEMS formatted inputs (step \circled{\footnotesize{B2}}) and the control of user appliances (step \circled{\footnotesize{B3}}).

\subsection{Integration of LLM} \label{integration}
To use the HEMS problem defined in section \ref{HEMS}, one needs to know the user's schedule (arrival and leaving time), its preferences (minimum and maximum temperatures, start and end date of simulation), and characteristics like the number of electric vehicles he owns (see Table \ref{tab:params} in the appendix). This is the job of an LLM-based agent that we will define in this section.

The interface needs to understand user stylometry, user dwellings specificities, user energy management preferences, user energy Home Energy Management System (HEMS) understanding. The user answers can contain vague information. Classical NLP models are not capable of performing on these wide ranges of complicated tasks or require huge efforts in fine-tuning, that is why using an LLM is required.

In sub flowchart B of Fig. \ref{fig:highlevelchart}, the LLM-agent interacts with the user (step \circled{B1}) and the HEMS (step \circled{B2}). The LLM agent must be able to communicate with both entities. To do so, we provide communication functions or tools that are all stored in a toolkit object. Each tool is described in the LLM agent instantiation prompt with the function of the tool, its inputs, and outputs. We want our LLM agent to be able to ask questions to a user based on a task, retrieve the wanted information from the user's answer, and store the retrieved parameters. To do so, we allow our LLM agent to use several tools such as $ask\_user$ and $store$ which will enable better communication between the user and the LLM agent and between the HEMS and the LLM agent. To properly use those tools, the LLM agent will have to generate function calls by providing the name of a tool and its input. The LLM agent will have to choose in the tool list to which it has access, by understanding the most appropriate tool to solve the problem it is currently working on. This method is known as function calling \cite{Schick2023ToolformerLM}.

We need our LLM agent to generate very precise texts for function calling, to do so we will provide examples which is a common practice. The few-shot prompting technique \cite{brown2020languagemodelsfewshotlearners} consists in adding one or several task-solving examples, giving more information to the LLM agent after the agent prompt template (see left of Fig. \ref{fig:flowchart} in the appendix). This will give more context to the LLM agent and diminish hallucination.

\subsection{ReAct} \label{react}
Many methods of prompt engineering have been developed to get the most out of LLMs in NLP tasks. In this study we have chosen to focus on ReAct \cite{yao2023react}. This method gives a ReAct prompt to the LLM to generate both reasoning traces and task-specific actions in an interleaved manner. This allows the system to perform dynamic reasoning to create, maintain, and adjust plans for acting while enabling interaction with external environments to incorporate additional information into the reasoning. The LLM agent will break down its tasks into smaller goals to increase its success rate (see right of Fig. \ref{fig:flowchart}).

In the agent prompt template (see Fig. \ref{fig:agentprompt} in section \ref{appendixagent} of the appendix) given to the LLM agent type $React+example$, we first describe the main goal of the LLM agent (line 1), the tools and their description (lines 1 and 5), and then the format rules it must follow for function calling (lines 3 to 13) and for ReAct (lines 15 to 27). After receiving a task, the LLM agent will have to rephrase it, then think about the first objective it must fulfill, and act to complete it by generating a correct JSON object that will be parsed to call a function. If the JSON blob is correct, the output of the function will be given to the LLM agent as an observation. The process continues until the LLM agent thinks that the task is completed.

\section{LLM user} \label{user}
For testing purposes, we have designed an LLM to operate in the place of a real user to gain testing time and model various behaviors. In the following paragraphs, we will refer to this LLM as the user. To simulate the diversity of LLM agent--user interaction, we set a high LLM temperature (0.8 in our tests) that allows for more diversity by increasing the weights of less likely LLM-generated words. The LLM user is given personal information that an LLM agent will try to retrieve by asking questions. The user prompt template is fairly simple and follows the same rules as the agent prompt template, such as goal description, context, and rules.

As in a real-life scenario, user behavior can be very different from person to person. We tried to model different users by giving them different prompts that will imply different answers to the LLM agent's questions. To design the user prompts, we used synonyms and changed the formats while also adding noisy information. We have defined three levels of difficulty for the LLM agent. In the easy mode, the user's answers are considered simple and to be exactly what the agent asked for (ex: date format). We define user precision as the quality of the user's answers. It measures how badly formatted the answers of the user are (see Fig \ref{fig:highlevelchart}). The user precision diminishes with the user difficulty mode. The user precision evaluation methodology is described in section \ref{userappendix} of the appendix.

\begin{table}[ht]
  \begin{center}
    \caption{Average cosine similarity score per difficulty level}
    \label{tab:precision}
    \begin{tabular}{c|c|c|c}
      \textbf{Difficulty} & Easy (E) & Medium (M) & Hard (H)\\
      \hline
      \textbf{Score} & 0.98 & 0.94 & 0.84
    \end{tabular}
  \end{center}
\end{table}

\section{Results} \label{results}
In this section, we present results demonstrating the performance of the proposed LLM-based HEMS interface. First, we explain and justify the settings used for the HEMS, the LLM user, and the LLM agent in our case studies. The experimental setup of the study is detailed in section \ref{resultsappendix} of the appendix.

We have chosen to make the LLM agent retrieve parameters of different types to check its generalization capability (see Tab. \ref{tab:params}). To test the added value of ReAct and the few-shot prompting, we developed three agent types: $Act$, $Act+example$, and $ReAct+example$. All three use function calling to interact with the user and the HEMS, so they can all "Act" by deciding to use a communication tool. To evaluate the importance of function calling, we developed agent type $Act$ that does not receive any example of how it should solve tasks. To evaluate the importance of ReAct, we implemented $Act+example$ agent type. It receives an example in its prompt of how it should solve tasks but does not follow the ReAct methodology which forces the agent to think and adapt to the results of its last action. An HEMS parametrization systems result is presented in Fig. \ref{fig:HEMS} in the appendix. It refers to a user-optimal energy management scenario. Based on our price data the user paid around 16£. If he had consumed the same amount of energy without taking into account energy price fluctuation, he would have paid around 31£ which is a 48\% energy cost reduction.

We have tested our interface with three different LLM versions of MistralAI's LLMs on 20 tests for each difficulty level \cite{wolf-etal-2020-transformers}. For notation simplicity, we respectively denote models OpenHermes-2.5-Mistral-7B, Mistral-7B-Instruct-v0.2, and Mistral-7B-Instruct-v0.3 as V1, V2, V3. Each LLM agent type was tested for all user difficulty modes Easy, Medium, and Hard (denoted as E, M, H). We want our LLM agent to store a value for all eight parameters at each test. To monitor the test's success, we will compute the accuracy of our interface. For each test, we will check if the parameters stored by the LLM agent are exactly the expected ones as these parameters are going to be used by the home energy management system that is inflexible: for example "Oxford" is different from "OXFORD" if you want to then use this city name to look up weather forecast data. As described in section \ref{user}, we instantiate the LLM user with a prompt template of a given difficulty and personal information. For each test, we will store the user's personal information and the parameter the LLM agent stored by himself and we will count this test as a success if the parameters are exactly the ones retrieved by the agent.

We have defined a home energy management system that simulates the electric house consumption of a person (section \ref{HEMS}). We can model different user behaviors with an LLM (section \ref{user}) and we can extract parameters with an LLM agent (section \ref{methodology}). For more detail on the parameter retrieval procedure, see section \ref{resultsappendix} of the appendix.

\subsection{Parameter retrieval effectiveness}
To better test the reliability of the interface, we need to run more tests to fully evaluate the strengths and limits of our work. Tab. \ref{tab:accuracy} shows each agent type's average accuracy per difficulty level. Overall, the V2-powered LLM agent achieved the best results (except in Easy mode). The accuracy of all LLMs is decreasing with rising difficulty, demonstrating the importance of user precision in extracting parameters. We observe that agent type $Act$ is performing worse than $Act+example$ which performs slightly worse than $ReAct+example$ (except for Model V3 in Hard and Medium). Thus, it highlights the significance of examples in prompts and the added value of ReAct compared to Act-only. 

\begin{table}[ht]
    \begin{center}
		\caption{Accuracy (\%) for different agent types and difficulty levels.}
		\label{tab:accuracy}
		\resizebox{0.47\textwidth}{!}{
        \begin{tabular}{c|c||c c c}
          \hline
          \textbf{LLM} & \textbf{Agent type} & \textbf{Easy (E)} & \textbf{Medium (M)} & \textbf{Hard (H)} \\
          \hline
          \multirow{2}{*}{V1} & Act & 70.6 & 63.1 & 53.1\\
          & Act+example & 85.6 & 74.4 & 65.6\\
          & ReAct+example & \textbf{97.5} & 84.4 & 73.1\\
          \hline
          \multirow{2}{*}{V2} & Act & 89.4 & 82.5 & 70.6\\
          & Act+example & 95 & 83.75 & 77.5 \\
          & ReAct+example & 96.9 & \textbf{87.5} & \textbf{78.8}\\
          \hline
          \multirow{2}{*}{V3} & Act & 88.1 & 83.1 & 74.4\\
          & Act+example & 89.4 & 78.8 & 68.1\\
          & ReAct+example & 91.9 & 86.3 & 69.4\\
          \hline
        \end{tabular}}
	\end{center}
\end{table}

With Fig. \ref{fig:results} in section \ref{resultsappendix} of the appendix, we see the significance of ReAct and few-shot prompting methods in our context which defines a faster and more accurate agent. V2 with $Act$ and V2 with $Act + examples$ are slow, and the proposed V2 with $React+example$ is faster and requires fewer questions. Even though agent type $Act+example$ has similar accuracy (see Tab. \ref{tab:accuracy}), this method requires more user questions than the $ReAct+example$ method in every difficulty level. As a reminder, an iteration is the process of asking the LLM agent to solve a task, i.e. to retrieve a parameter. Sometimes, the LLM agent fails to store a value, so it needs more than eight iterations to find all eight parameters. $Act$ method is by far the longest method in terms of iteration and, therefore also in duration. We observe that $ReAct+example$ performs better than $Act+example$ in all three difficulty levels. Fig. \ref{fig:results} also shows that model V1 is far slower than models V3 and V2 and requires more iterations and questions.

\section{Conclusion} \label{conclusion}
With this study, we have moved toward the democratization of home energy management systems. We have developed a representative use case of a single home-owner interaction with an LLM agent that enables a person without technical knowledge to specify key parameters for home energy optimisation. The $ReAct+example$ LLM agent has shown remarkable adaptability by being able to retrieve parameters with different levels of user precision with higher retrieval accuracy and a smaller number of questions than with other methods.

Several leads can be exploited to correct the current limitations of our interface. Inferring more technical parameters for a more accurate HEMS like window-to-wall ratio or house orientation from a user non-technical description, user photos and house plan would make our approach more relevant as they can be easily obtained from a user and can provide much information the user may not be able to render to the LLM agent directly. Such tasks may require a multi-modal LLM. LLMs can provide extended user interactions. LLM-based user interactive systems can benefit from API integration like web searches or classical NLP approaches. We think that NLP algorithms and LLMs are complementary and could be combined in future work in a faster and more efficient HEMS parameterization approach. User testing is essential to validate the usability and effectiveness of the LLM-based HEMS. The LLM-user introduced in this study models a wide variety of user behavior, pushing the LLM-agent to its limits and thus providing a solid testing approach. However, while our proposed LLM-based approach enables efficient testing, future efforts should incorporate pilot studies in diverse real households to gather feedback, evaluate performance, and refine the system for broader and more practical adoption. In future works, a more reliable LLM-based system should be tested with a large corpora of real independent users to fully evaluate the capabilities of the system. The test results will provide new objectives and highlight key points that will need to be addressed.

We might also want our LLM agent to take action based on its understanding of the user behavior, to be able to propose more adapted consumption strategies and explain them to the user.
Being able to answer user questions, like "What should I do to diminish my electric bills ?" or basic ones on the algorithm documentation, would be a plus. One can imagine a multi-modal interface that could be combined with Retrieval-Augmented Generation to retrieve information from a range of data sources. In the case where a parameter has not been retrieved like house thermal characteristics, we can imagine an interface where the LLM agent could cross reference incomplete textual information from the user with other sources, such as online information about similar buildings or sensor readings.

\begin{acks}
This work was supported by EDF Energy and the Maison Française d’Oxford. I want to express my deepest appreciation to everyone involved in the Department of Engineering Science of Oxford - EDF Energy scheme that proposed the research project I have worked on and have presented in this article. EDF Energy's generous contributions have made it possible to conduct this study and have significantly contributed to its successful completion.
\end{acks}

\bibliographystyle{ACM-Reference-Format}
\bibliography{references}


\begin{thebibliography}{27}


\ifx \showCODEN    \undefined \def \showCODEN     #1{\unskip}     \fi
\ifx \showDOI      \undefined \def \showDOI       #1{#1}\fi
\ifx \showISBNx    \undefined \def \showISBNx     #1{\unskip}     \fi
\ifx \showISBNxiii \undefined \def \showISBNxiii  #1{\unskip}     \fi
\ifx \showISSN     \undefined \def \showISSN      #1{\unskip}     \fi
\ifx \showLCCN     \undefined \def \showLCCN      #1{\unskip}     \fi
\ifx \shownote     \undefined \def \shownote      #1{#1}          \fi
\ifx \showarticletitle \undefined \def \showarticletitle #1{#1}   \fi
\ifx \showURL      \undefined \def \showURL       {\relax}        \fi
\providecommand\bibfield[2]{#2}
\providecommand\bibinfo[2]{#2}
\providecommand\natexlab[1]{#1}
\providecommand\showeprint[2][]{arXiv:#2}

\bibitem[Agency(2023)]%
        {europa}
\bibfield{author}{\bibinfo{person}{European~Environment Agency}.} \bibinfo{year}{2023}\natexlab{}.
\newblock \bibinfo{title}{eea.europa.eu}.
\newblock \bibinfo{howpublished}{\url{https://www.eea.europa.eu/publications/flexibility-solutions-to-support/at_download/file}}.
\newblock
\newblock
\shownote{[Accessed 19-07-2024]}.


\bibitem[Bonadia et~al\mbox{.}(2023)]%
        {10251972}
\bibfield{author}{\bibinfo{person}{Rodrigo~S. Bonadia}, \bibinfo{person}{Fernanda C.~L. Trindade}, \bibinfo{person}{Walmir Freitas}, {and} \bibinfo{person}{Bala Venkatesh}.} \bibinfo{year}{2023}\natexlab{}.
\newblock \showarticletitle{On the Potential of ChatGPT to Generate Distribution Systems for Load Flow Studies Using OpenDSS}.
\newblock \bibinfo{journal}{\emph{IEEE Transactions on Power Systems}} \bibinfo{volume}{38}, \bibinfo{number}{6} (\bibinfo{year}{2023}), \bibinfo{pages}{5965--5968}.
\newblock
\urldef\tempurl%
\url{https://doi.org/10.1109/TPWRS.2023.3315543}
\showDOI{\tempurl}


\bibitem[Brown et~al\mbox{.}(2020)]%
        {brown2020languagemodelsfewshotlearners}
\bibfield{author}{\bibinfo{person}{Tom~B. Brown}, \bibinfo{person}{Benjamin Mann}, \bibinfo{person}{Nick Ryder}, \bibinfo{person}{Melanie Subbiah}, \bibinfo{person}{Jared Kaplan}, \bibinfo{person}{Prafulla Dhariwal}, \bibinfo{person}{Arvind Neelakantan}, \bibinfo{person}{Pranav Shyam}, \bibinfo{person}{Girish Sastry}, \bibinfo{person}{Amanda Askell}, \bibinfo{person}{Sandhini Agarwal}, \bibinfo{person}{Ariel Herbert-Voss}, \bibinfo{person}{Gretchen Krueger}, \bibinfo{person}{Tom Henighan}, \bibinfo{person}{Rewon Child}, \bibinfo{person}{Aditya Ramesh}, \bibinfo{person}{Daniel~M. Ziegler}, \bibinfo{person}{Jeffrey Wu}, \bibinfo{person}{Clemens Winter}, \bibinfo{person}{Christopher Hesse}, \bibinfo{person}{Mark Chen}, \bibinfo{person}{Eric Sigler}, \bibinfo{person}{Mateusz Litwin}, \bibinfo{person}{Scott Gray}, \bibinfo{person}{Benjamin Chess}, \bibinfo{person}{Jack Clark}, \bibinfo{person}{Christopher Berner}, \bibinfo{person}{Sam McCandlish}, \bibinfo{person}{Alec Radford}, \bibinfo{person}{Ilya Sutskever},
  {and} \bibinfo{person}{Dario Amodei}.} \bibinfo{year}{2020}\natexlab{}.
\newblock \bibinfo{title}{Language Models are Few-Shot Learners}.
\newblock
\newblock
\showeprint[arxiv]{2005.14165}~[cs.CL]
\urldef\tempurl%
\url{https://arxiv.org/abs/2005.14165}
\showURL{%
\tempurl}


\bibitem[CLNR(2024)]%
        {networkrevolutionHomepageCustomerLed}
\bibfield{author}{\bibinfo{person}{CLNR}.} \bibinfo{year}{2024}\natexlab{}.
\newblock \bibinfo{title}{{H}omepage - {C}ustomer-{L}ed {N}etwork {R}evolution --- networkrevolution.co.uk}.
\newblock \bibinfo{howpublished}{\url{http://www.networkrevolution.co.uk/}}.
\newblock
\newblock
\shownote{[Accessed 13-08-2024]}.


\bibitem[EsaLaboratory(2024)]%
        {LLM}
\bibfield{author}{\bibinfo{person}{EsaLaboratory}.} \bibinfo{year}{2024}\natexlab{}.
\newblock \bibinfo{title}{{LLM}}.
\newblock
\newblock
\urldef\tempurl%
\url{https://github.com/EsaLaboratory/llm}
\showURL{%
\tempurl}


\bibitem[fei Chen et~al\mbox{.}(2020)]%
        {CHEN2020101616}
\bibfield{author}{\bibinfo{person}{Chien fei Chen}, \bibinfo{person}{Xiaojing Xu}, \bibinfo{person}{Jacqueline Adams}, \bibinfo{person}{James Brannon}, \bibinfo{person}{Fangxing Li}, {and} \bibinfo{person}{Allen Walzem}.} \bibinfo{year}{2020}\natexlab{}.
\newblock \showarticletitle{When East meets West: Understanding residents’ home energy management system adoption intention and willingness to pay in Japan and the United States}.
\newblock \bibinfo{journal}{\emph{Energy Research Social Science}}  \bibinfo{volume}{69} (\bibinfo{year}{2020}), \bibinfo{pages}{101616}.
\newblock
\showISSN{2214-6296}
\urldef\tempurl%
\url{https://doi.org/10.1016/j.erss.2020.101616}
\showDOI{\tempurl}


\bibitem[Giudici et~al\mbox{.}(2024)]%
        {designs8030043}
\bibfield{author}{\bibinfo{person}{Mathyas Giudici}, \bibinfo{person}{Luca Padalino}, \bibinfo{person}{Giovanni Paolino}, \bibinfo{person}{Ilaria Paratici}, \bibinfo{person}{Alexandru~Ionut Pascu}, {and} \bibinfo{person}{Franca Garzotto}.} \bibinfo{year}{2024}\natexlab{}.
\newblock \showarticletitle{Designing Home Automation Routines Using an LLM-Based Chatbot}.
\newblock \bibinfo{journal}{\emph{Designs}} \bibinfo{volume}{8}, \bibinfo{number}{3} (\bibinfo{year}{2024}).
\newblock
\showISSN{2411-9660}
\urldef\tempurl%
\url{https://doi.org/10.3390/designs8030043}
\showDOI{\tempurl}


\bibitem[Han et~al\mbox{.}(2011)]%
        {5973168}
\bibfield{author}{\bibinfo{person}{Jinsoo Han}, \bibinfo{person}{Chang-Sic Choi}, \bibinfo{person}{Wan-Ki Park}, {and} \bibinfo{person}{Ilwoo Lee}.} \bibinfo{year}{2011}\natexlab{}.
\newblock \showarticletitle{Green Home Energy Management System through comparison of energy usage between the same kinds of home appliances}. In \bibinfo{booktitle}{\emph{2011 IEEE 15th International Symposium on Consumer Electronics (ISCE)}}. \bibinfo{publisher}{IEEE}, \bibinfo{address}{Singapore}, \bibinfo{pages}{1--4}.
\newblock
\urldef\tempurl%
\url{https://doi.org/10.1109/ISCE.2011.5973168}
\showDOI{\tempurl}


\bibitem[Hao et~al\mbox{.}(2015)]%
        {6832599}
\bibfield{author}{\bibinfo{person}{He Hao}, \bibinfo{person}{Borhan~M. Sanandaji}, \bibinfo{person}{Kameshwar Poolla}, {and} \bibinfo{person}{Tyrone~L. Vincent}.} \bibinfo{year}{2015}\natexlab{}.
\newblock \showarticletitle{Aggregate Flexibility of Thermostatically Controlled Loads}.
\newblock \bibinfo{journal}{\emph{IEEE Transactions on Power Systems}} \bibinfo{volume}{30}, \bibinfo{number}{1} (\bibinfo{year}{2015}), \bibinfo{pages}{189--198}.
\newblock
\urldef\tempurl%
\url{https://doi.org/10.1109/TPWRS.2014.2328865}
\showDOI{\tempurl}


\bibitem[Huang et~al\mbox{.}(2023)]%
        {huang2023largefoundationmodelspower}
\bibfield{author}{\bibinfo{person}{Chenghao Huang}, \bibinfo{person}{Siyang Li}, \bibinfo{person}{Ruohong Liu}, \bibinfo{person}{Hao Wang}, {and} \bibinfo{person}{Yize Chen}.} \bibinfo{year}{2023}\natexlab{}.
\newblock \bibinfo{title}{Large Foundation Models for Power Systems}.
\newblock
\newblock
\showeprint[arxiv]{2312.07044}~[eess.SY]
\urldef\tempurl%
\url{https://arxiv.org/abs/2312.07044}
\showURL{%
\tempurl}


\bibitem[Jia et~al\mbox{.}(2024)]%
        {jia2024enablinglargelanguagemodels}
\bibfield{author}{\bibinfo{person}{Mengshuo Jia}, \bibinfo{person}{Zeyu Cui}, {and} \bibinfo{person}{Gabriela Hug}.} \bibinfo{year}{2024}\natexlab{}.
\newblock \bibinfo{title}{Enabling Large Language Models to Perform Power System Simulations with Previously Unseen Tools: A Case of Daline}.
\newblock
\newblock
\showeprint[arxiv]{2406.17215}~[eess.SY]
\urldef\tempurl%
\url{https://arxiv.org/abs/2406.17215}
\showURL{%
\tempurl}


\bibitem[Lee and Cheng(2016)]%
        {LEE2016760}
\bibfield{author}{\bibinfo{person}{Dasheng Lee} {and} \bibinfo{person}{Chin-Chi Cheng}.} \bibinfo{year}{2016}\natexlab{}.
\newblock \showarticletitle{Energy savings by energy management systems: A review}.
\newblock \bibinfo{journal}{\emph{Renewable and Sustainable Energy Reviews}}  \bibinfo{volume}{56} (\bibinfo{year}{2016}), \bibinfo{pages}{760--777}.
\newblock
\showISSN{1364-0321}
\urldef\tempurl%
\url{https://doi.org/10.1016/j.rser.2015.11.067}
\showDOI{\tempurl}


\bibitem[Majumder et~al\mbox{.}(2024)]%
        {MAJUMDER20241544}
\bibfield{author}{\bibinfo{person}{Subir Majumder}, \bibinfo{person}{Lin Dong}, \bibinfo{person}{Fatemeh Doudi}, \bibinfo{person}{Yuting Cai}, \bibinfo{person}{Chao Tian}, \bibinfo{person}{Dileep Kalathil}, \bibinfo{person}{Kevin Ding}, \bibinfo{person}{Anupam~A. Thatte}, \bibinfo{person}{Na Li}, {and} \bibinfo{person}{Le Xie}.} \bibinfo{year}{2024}\natexlab{}.
\newblock \showarticletitle{Exploring the capabilities and limitations of large language models in the electric energy sector}.
\newblock \bibinfo{journal}{\emph{Joule}} \bibinfo{volume}{8}, \bibinfo{number}{6} (\bibinfo{year}{2024}), \bibinfo{pages}{1544--1549}.
\newblock
\showISSN{2542-4351}
\urldef\tempurl%
\url{https://doi.org/10.1016/j.joule.2024.05.009}
\showDOI{\tempurl}


\bibitem[Matsuyama(2014)]%
        {Matsuyama2014}
\bibfield{author}{\bibinfo{person}{Takashi Matsuyama}.} \bibinfo{year}{2014}\natexlab{}.
\newblock \bibinfo{booktitle}{\emph{i-Energy: Smart Demand-Side Energy Management}}.
\newblock \bibinfo{publisher}{Springer London}, \bibinfo{address}{London}, \bibinfo{pages}{141--163}.
\newblock
\showISBNx{978-1-4471-6281-0}
\urldef\tempurl%
\url{https://doi.org/10.1007/978-1-4471-6281-0_8}
\showDOI{\tempurl}


\bibitem[Morstyn et~al\mbox{.}(2020)]%
        {MORSTYN2020115397}
\bibfield{author}{\bibinfo{person}{Thomas Morstyn}, \bibinfo{person}{Katherine~A. Collett}, \bibinfo{person}{Avinash Vijay}, \bibinfo{person}{Matthew Deakin}, \bibinfo{person}{Scot Wheeler}, \bibinfo{person}{Sivapriya~M. Bhagavathy}, \bibinfo{person}{Filiberto Fele}, {and} \bibinfo{person}{Malcolm~D. McCulloch}.} \bibinfo{year}{2020}\natexlab{}.
\newblock \showarticletitle{OPEN: An open-source platform for developing smart local energy system applications}.
\newblock \bibinfo{journal}{\emph{Applied Energy}}  \bibinfo{volume}{275} (\bibinfo{year}{2020}), \bibinfo{pages}{115397}.
\newblock
\showISSN{0306-2619}
\urldef\tempurl%
\url{https://doi.org/10.1016/j.apenergy.2020.115397}
\showDOI{\tempurl}


\bibitem[Pfenninger and Staffell(2016)]%
        {PFENNINGER20161251}
\bibfield{author}{\bibinfo{person}{Stefan Pfenninger} {and} \bibinfo{person}{Iain Staffell}.} \bibinfo{year}{2016}\natexlab{}.
\newblock \showarticletitle{Long-term patterns of European PV output using 30 years of validated hourly reanalysis and satellite data}.
\newblock \bibinfo{journal}{\emph{Energy}}  \bibinfo{volume}{114} (\bibinfo{year}{2016}), \bibinfo{pages}{1251--1265}.
\newblock
\showISSN{0360-5442}
\urldef\tempurl%
\url{https://doi.org/10.1016/j.energy.2016.08.060}
\showDOI{\tempurl}


\bibitem[Rivkin et~al\mbox{.}(2024)]%
        {10729865}
\bibfield{author}{\bibinfo{person}{Dmitriy Rivkin}, \bibinfo{person}{Francois Hogan}, \bibinfo{person}{Amal Feriani}, \bibinfo{person}{Abhisek Konar}, \bibinfo{person}{Adam Sigal}, \bibinfo{person}{Xue Liu}, {and} \bibinfo{person}{Gregory Dudek}.} \bibinfo{year}{2024}\natexlab{}.
\newblock \showarticletitle{AIoT Smart Home via Autonomous LLM Agents}.
\newblock \bibinfo{journal}{\emph{IEEE Internet of Things Journal}} (\bibinfo{year}{2024}), \bibinfo{pages}{1--1}.
\newblock
\urldef\tempurl%
\url{https://doi.org/10.1109/JIOT.2024.3471904}
\showDOI{\tempurl}


\bibitem[Schick et~al\mbox{.}(2023)]%
        {Schick2023ToolformerLM}
\bibfield{author}{\bibinfo{person}{Timo Schick}, \bibinfo{person}{Jane Dwivedi-Yu}, \bibinfo{person}{Roberto Dess{\`i}}, \bibinfo{person}{Roberta Raileanu}, \bibinfo{person}{Maria Lomeli}, \bibinfo{person}{Luke Zettlemoyer}, \bibinfo{person}{Nicola Cancedda}, {and} \bibinfo{person}{Thomas Scialom}.} \bibinfo{year}{2023}\natexlab{}.
\newblock \showarticletitle{Toolformer: Language Models Can Teach Themselves to Use Tools}.
\newblock \bibinfo{journal}{\emph{ArXiv}}  \bibinfo{volume}{abs/2302.04761} (\bibinfo{year}{2023}).
\newblock
\urldef\tempurl%
\url{https://api.semanticscholar.org/CorpusID:256697342}
\showURL{%
\tempurl}


\bibitem[Staffell and Pfenninger(2016)]%
        {STAFFELL20161224}
\bibfield{author}{\bibinfo{person}{Iain Staffell} {and} \bibinfo{person}{Stefan Pfenninger}.} \bibinfo{year}{2016}\natexlab{}.
\newblock \showarticletitle{Using bias-corrected reanalysis to simulate current and future wind power output}.
\newblock \bibinfo{journal}{\emph{Energy}}  \bibinfo{volume}{114} (\bibinfo{year}{2016}), \bibinfo{pages}{1224--1239}.
\newblock
\showISSN{0360-5442}
\urldef\tempurl%
\url{https://doi.org/10.1016/j.energy.2016.08.068}
\showDOI{\tempurl}


\bibitem[Stavrakas and Flamos(2020)]%
        {STAVRAKAS2020112339}
\bibfield{author}{\bibinfo{person}{Vassilis Stavrakas} {and} \bibinfo{person}{Alexandros Flamos}.} \bibinfo{year}{2020}\natexlab{}.
\newblock \showarticletitle{A modular high-resolution demand-side management model to quantify benefits of demand-flexibility in the residential sector}.
\newblock \bibinfo{journal}{\emph{Energy Conversion and Management}}  \bibinfo{volume}{205} (\bibinfo{year}{2020}), \bibinfo{pages}{112339}.
\newblock
\showISSN{0196-8904}
\urldef\tempurl%
\url{https://doi.org/10.1016/j.enconman.2019.112339}
\showDOI{\tempurl}


\bibitem[Toshevska et~al\mbox{.}(2020)]%
        {Toshevska_2020}
\bibfield{author}{\bibinfo{person}{Martina Toshevska}, \bibinfo{person}{Frosina Stojanovska}, {and} \bibinfo{person}{Jovan Kalajdjieski}.} \bibinfo{year}{2020}\natexlab{}.
\newblock \showarticletitle{Comparative Analysis of Word Embeddings for Capturing Word Similarities}. In \bibinfo{booktitle}{\emph{6th International Conference on Natural Language Processing (NATP 2020)}} \emph{(\bibinfo{series}{NATP 2020})}. \bibinfo{publisher}{Aircc Publishing Corporation}, \bibinfo{address}{Copenhagen, Denmark}.
\newblock
\urldef\tempurl%
\url{https://doi.org/10.5121/csit.2020.100402}
\showDOI{\tempurl}


\bibitem[Verschae et~al\mbox{.}(2016)]%
        {en9070562}
\bibfield{author}{\bibinfo{person}{Rodrigo Verschae}, \bibinfo{person}{Takekazu Kato}, {and} \bibinfo{person}{Takashi Matsuyama}.} \bibinfo{year}{2016}\natexlab{}.
\newblock \showarticletitle{Energy Management in Prosumer Communities: A Coordinated Approach}.
\newblock \bibinfo{journal}{\emph{Energies}} \bibinfo{volume}{9}, \bibinfo{number}{7} (\bibinfo{year}{2016}).
\newblock
\showISSN{1996-1073}
\urldef\tempurl%
\url{https://doi.org/10.3390/en9070562}
\showDOI{\tempurl}


\bibitem[Wang et~al\mbox{.}(2020)]%
        {wang2020minilmdeepselfattentiondistillation}
\bibfield{author}{\bibinfo{person}{Wenhui Wang}, \bibinfo{person}{Furu Wei}, \bibinfo{person}{Li Dong}, \bibinfo{person}{Hangbo Bao}, \bibinfo{person}{Nan Yang}, {and} \bibinfo{person}{Ming Zhou}.} \bibinfo{year}{2020}\natexlab{}.
\newblock \bibinfo{title}{MiniLM: Deep Self-Attention Distillation for Task-Agnostic Compression of Pre-Trained Transformers}.
\newblock
\newblock
\showeprint[arxiv]{2002.10957}~[cs.CL]
\urldef\tempurl%
\url{https://arxiv.org/abs/2002.10957}
\showURL{%
\tempurl}


\bibitem[Wolf et~al\mbox{.}(2020)]%
        {wolf-etal-2020-transformers}
\bibfield{author}{\bibinfo{person}{Thomas Wolf}, \bibinfo{person}{Lysandre Debut}, \bibinfo{person}{Victor Sanh}, \bibinfo{person}{Julien Chaumond}, \bibinfo{person}{Clement Delangue}, \bibinfo{person}{Anthony Moi}, \bibinfo{person}{Pierric Cistac}, \bibinfo{person}{Tim Rault}, \bibinfo{person}{Rémi Louf}, \bibinfo{person}{Morgan Funtowicz}, \bibinfo{person}{Joe Davison}, \bibinfo{person}{Sam Shleifer}, \bibinfo{person}{Patrick von Platen}, \bibinfo{person}{Clara Ma}, \bibinfo{person}{Yacine Jernite}, \bibinfo{person}{Julien Plu}, \bibinfo{person}{Canwen Xu}, \bibinfo{person}{Teven~Le Scao}, \bibinfo{person}{Sylvain Gugger}, \bibinfo{person}{Mariama Drame}, \bibinfo{person}{Quentin Lhoest}, {and} \bibinfo{person}{Alexander~M. Rush}.} \bibinfo{year}{2020}\natexlab{}.
\newblock \showarticletitle{Transformers: State-of-the-Art Natural Language Processing}. In \bibinfo{booktitle}{\emph{Proceedings of the 2020 Conference on Empirical Methods in Natural Language Processing: System Demonstrations}}. \bibinfo{publisher}{Association for Computational Linguistics}, \bibinfo{address}{Online}, \bibinfo{pages}{38--45}.
\newblock
\urldef\tempurl%
\url{https://www.aclweb.org/anthology/2020.emnlp-demos.6}
\showURL{%
\tempurl}


\bibitem[Yao et~al\mbox{.}(2023)]%
        {yao2023react}
\bibfield{author}{\bibinfo{person}{Shunyu Yao}, \bibinfo{person}{Jeffrey Zhao}, \bibinfo{person}{Dian Yu}, \bibinfo{person}{Nan Du}, \bibinfo{person}{Izhak Shafran}, \bibinfo{person}{Karthik Narasimhan}, {and} \bibinfo{person}{Yuan Cao}.} \bibinfo{year}{2023}\natexlab{}.
\newblock \bibinfo{title}{ReAct: Synergizing Reasoning and Acting in Language Models}.
\newblock
\newblock
\showeprint[arxiv]{2210.03629}~[cs.CL]


\bibitem[Zeid et~al\mbox{.}(2024)]%
        {zeid2024predictingdclinkcapacitorcurrent}
\bibfield{author}{\bibinfo{person}{Mohamed Zeid}, \bibinfo{person}{Subir Majumder}, \bibinfo{person}{Hasan Ibrahim}, \bibinfo{person}{Prasad Enjeti}, \bibinfo{person}{Le Xie}, {and} \bibinfo{person}{Chao Tian}.} \bibinfo{year}{2024}\natexlab{}.
\newblock \bibinfo{title}{Predicting DC-Link Capacitor Current Ripple in AC-DC Rectifier Circuits Using Fine-Tuned Large Language Models}.
\newblock
\newblock
\showeprint[arxiv]{2407.01724}~[eess.SY]
\urldef\tempurl%
\url{https://arxiv.org/abs/2407.01724}
\showURL{%
\tempurl}


\bibitem[Zhou et~al\mbox{.}(2016)]%
        {ZHOU201630}
\bibfield{author}{\bibinfo{person}{Bin Zhou}, \bibinfo{person}{Wentao Li}, \bibinfo{person}{Ka~Wing Chan}, \bibinfo{person}{Yijia Cao}, \bibinfo{person}{Yonghong Kuang}, \bibinfo{person}{Xi Liu}, {and} \bibinfo{person}{Xiong Wang}.} \bibinfo{year}{2016}\natexlab{}.
\newblock \showarticletitle{Smart home energy management systems: Concept, configurations, and scheduling strategies}.
\newblock \bibinfo{journal}{\emph{Renewable and Sustainable Energy Reviews}}  \bibinfo{volume}{61} (\bibinfo{year}{2016}), \bibinfo{pages}{30--40}.
\newblock
\showISSN{1364-0321}
\urldef\tempurl%
\url{https://doi.org/10.1016/j.rser.2016.03.047}
\showDOI{\tempurl}


\end{thebibliography}

\appendix
\section{Additional resources} \label{resources}

\subsection{HEMS} \label{hemsappendix}

Here we present in detail the model behind the HEMS of the study with Eq. \ref{eq:1}. The solar-generated electricity is directly consumed by user appliances, and if solar panels produce more energy than needed, the surplus is sold as highlighted in (Eq. \ref{eq:price}). Prices depend on the user's energy provider. The function the HEMS minimizes is defined in (Eq. \ref{eq:min}) and its variables are described in Tab. \ref{tab:variable}. We define $p_t(t)$ the total power the user is consuming at time t in (Eq. 
 \ref{eq:P_tot}). We did not consider V2G (the process of feeding the energy stored in an electric vehicle’s battery back into the National Grid), which is why in (Eq. \ref{eq:EV_ineq}) $P_{EV}$ is always positive. (Eq. \ref{eq:ev}) is the EV charging formula. (Eq. \ref{eq:e0}) and (Eq. \ref{eq:E_end}) are respectively the start and end condition constraints. 
The temperature evolution formulas and parameters in (Eq. \ref{eq:temp}) and (Eq. \ref{eq:param}) are taken from \cite{6832599}.

\begin{subequations}
	\begin{align}
		&\min_{p_{heat}, p_{EV}} \quad \sum_{t=1}^{T}(p_{t}(t) - P_{s}(t))\pi(t)\label{eq:min}\\
		&\textrm{s.t.} \quad \forall t=1,...,T\\ &p_{t}(t)=p_{heat}(t)+p_{EV}(t)+P_{other}(t) \label{eq:P_tot}\\
        &\pi(t) = \pi^e(t)\mathds{1}_{p_{t}(t) \geq P_{s}(t)}(t) + \pi^s(t)\mathds{1}_{p_{t}(t) < P_{s}(t)}(t)\label{eq:price}\\
		&P_{EV\_max}(t) \geq p_{EV}(t) \geq P_{EV\_min}(t)\label{eq:EV_ineq}\\
		&E_{EV}(t+dt) = E_{EV}(t) + p_{EV}(t)dt\label{eq:ev}\\ 
		&E_{EV}(t_{arrival}) = E_{init}\label{eq:e0}\\
		&E_{EV}(t_{departure}) = E_{fully\_charged}\label{eq:E_end}\\
		&\begin{aligned}
        T_{house}(t+dt) = &dt (\beta p_{heat}(t) + \alpha (T_{ext}(t) - T_{house}(t)))\\
        &+ T_{house}(t)
        \end{aligned}\label{eq:temp}\\
        &\alpha = \frac{1}{R_{th}C_{th}}, \quad \beta = \frac{\eta}{C_{th}}, \quad T_{max} \geq T_{house}(t) \geq T_{min}\label{eq:param}
	\end{align}
	\label{eq:1}
\end{subequations}

\begin{table}[ht]
  \begin{center}
    \caption{Variable description}
    \label{tab:variable}
    \begin{tabular}{c|c|c}
      \textbf{Variable} & \textbf{Description} & \textbf{Unit}\\
      \hline
      $p_{t}$ & total power consumption & kW\\
      $p_{heat}$ & heating power & kW\\
      $p_{EV}$ & EV(s) charging power & kW\\
      $P_{other}$ & power consumed by other appliances & kW\\
      $P_{s}$ & solar-generated power & kW\\
      $\pi^e$ & consumption electricity price & £/kWh\\
      $\pi^s$ & feed-in price of solar energy & £/kWh\\
      $E_{EV}$ & EV energy & kWh\\
      $T_{house}$ & house temperature & \degree C\\
      $T_{ext}$ & outdoor temperature & \degree C\\
      $C_{th}$ & thermal capacitance & kWh/ \degree C\\
      $R_{th}$ & thermal resistance & \degree C/kW\\
      $\eta$ & coefficient of performance & 1\\
    \end{tabular}
  \end{center}
\end{table}

Based on our HEMS in (Eq. \ref{eq:1}), Tab. \ref{tab:params} describes all the parameters that are required to define the HEMS's optimization problem and obtain the optimal consumption energy schedule. For example, with the user's location and the simulation period we can extract the outside temperature data and solar power generation \cite{PFENNINGER20161251} \cite{STAFFELL20161224}. There are many parameters of different types (dates, integers, floats, and strings). Each input needs to be perfectly formatted to be then used by the nonflexible HEMS in its optimization process. For a user without experience in HEMS or for the elderly, it is important to provide an easy-to-use interface while interacting with HEMS. An autofill interface could easily gather simple information but it can lack clarity for someone without experience. For example, a user may want to express his preferences in a detailed sentence that a single input cannot fully represent. Some parameters' names can also be confusing and too long parameter documentation can make the user puzzled. Ideally, a user should be able to precisely parametrize an HEMS with ease without being overwhelmed by the complexity of the required information. Parameters of Tab. \ref{tab:params} symbolize the information that can be expected from the user for the HEMS. Being able to easily retrieve them is key to addressing the complexity of HEMS parametrization.

\begin{table}[ht]
	\centering
	\caption{User parameters description}
	\resizebox{0.47\textwidth}{!}{\begin{tabular}{|c|p{3cm}|c|}
		\hline
		name & description & type\\
		\hline
		\hline
		date start, date end & start and end date of simulation & python datetime\\
		\hline
		EV & number of electric vehicles owned by the user & int\\
		\hline
		city & the city where the user lives in & string\\
		\hline
		EV start time, EV end time & user arrival time and leaving time & python time\\
		\hline		
		Tmin, Tmax & \raggedright minimum and maximum comfort house temperature & float\\
		\hline
	\end{tabular}}
	\label{tab:params}
\end{table}

\subsection{ReAct} \label{reactappendix}

\begin{figure*}[ht!]
  \includegraphics[width=\textwidth,height=5.8cm]{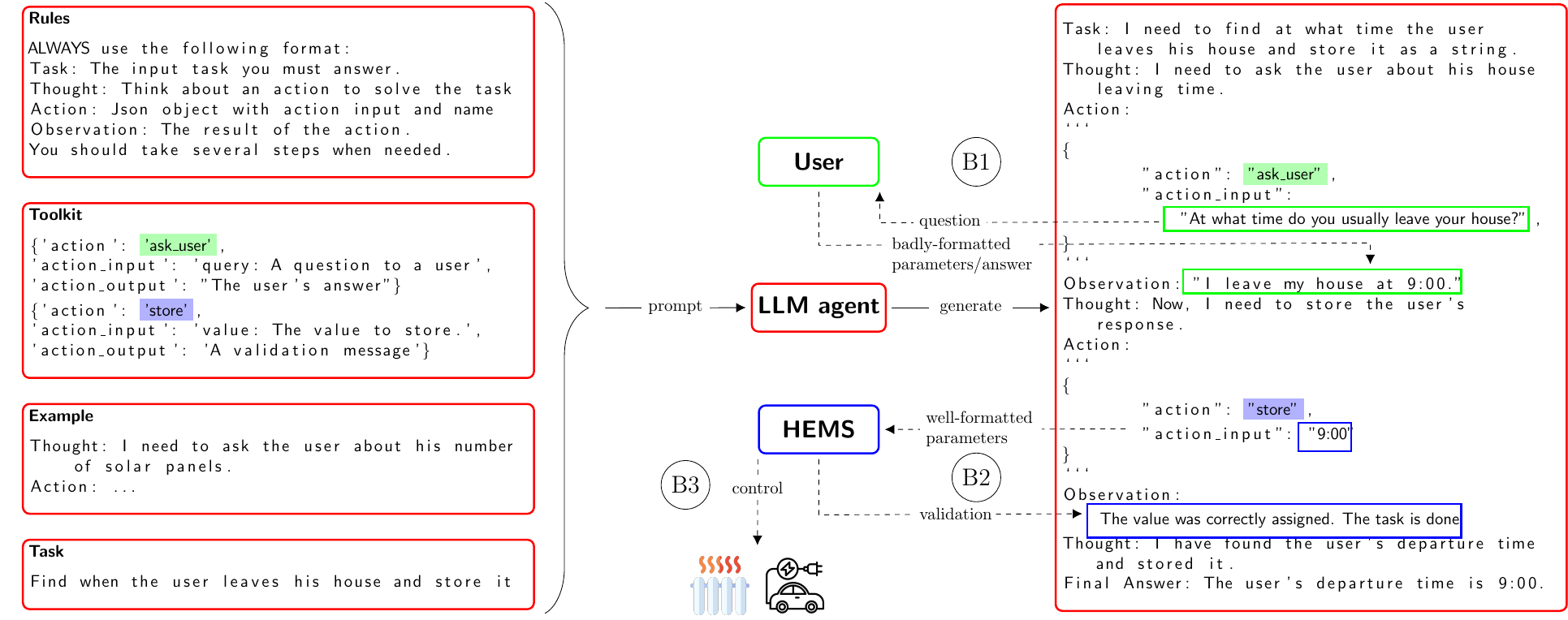}
  \caption[centerfirst]{Interface flow diagram of React retrieval \par \small \normalfont The \textbf{LLM agent} first receives a prompt which is divided into four parts: \textbf{Rules} that will specify the format it must follow, a \textbf{Toolkit} containing the communication tools the LLM agent can use when acting, an \textbf{Example} to give a hint to the agent on the expected work, and the actual \textbf{Task} it must solve. Step \circled{\footnotesize{B1}} is performed by calling the function \colorbox{green!30}{"ask\_user"} with the user question as input, it outputs the user's answer. In Step \circled{\footnotesize{B2}}, the \textbf{LLM agent} sends a formatted parameter to the \textbf{HEMS} by calling the function \colorbox{blue!30}{"store"}. This function returns a validation message when the parameter has the correct format. When all parameters are found, the \textbf{HEMS} can control the user's appliances in step \circled{\footnotesize{B3}}.}
  \Description{Fully described in the caption.}
  \label{fig:flowchart}
\end{figure*}

In this section, we detail the different components behind the ReAct agent of our HEMS parametrization system. The interface flowchart diagram shown in Fig. \ref{fig:flowchart} describes with more detail a parameter retrieval of agent type $React+example$ which is symbolized by steps \circled{B1} and \circled{B2} in Fig. \ref{fig:highlevelchart}. The LLM agent has to find each one of the user parameters thanks to the user's answers to the LLM agent's questions.

As shown in Algorithm \ref{alg:react}, the React algorithm of our interface starts by initiating the LLM agent with a prompt template and a task (line $2$). The algorithm stops when the maximum number of iterations is reached or when the LLM agent has found and stored the desired parameter that can be checked with the $is\_done$ function (line $3$). The LLM agent will generate texts containing the Thought/Action/Observation format (line $5$). The generated response is parsed. If the format is respected, the parser will output the result of the function called in the LLM-generated JSON (line $6$); otherwise, the parser will output an error message. In each case, we concatenate the newly generated text and the parser's output into the agent prompt (line $7$). Upon completion, the algorithm returns the stored value of the LLM in case of success.

\begin{algorithm}[ht!]
\caption{React LLM agent parameter retrieval algorithm} \label{alg:react}
\begin{algorithmic}[1]
\STATE $iter \gets 0$
\STATE $agent\_prompt \gets init\_prompt(task)$
\WHILE{\NOT  $is\_done()$ \AND $n\_iter \geq iter$}
    \STATE $iter \gets iter + 1$
    \STATE $response \gets generate(agent, agent\_prompt)$
    \STATE $output \gets parser(response)$
    \STATE $agent\_prompt \gets agent\_prompt + response + output$ 
\ENDWHILE
\RETURN $get\_result(agent\_prompt)$
\end{algorithmic}
\end{algorithm}

Algorithm \ref{alg:pipeline} describes the interface in detail. A toolkit object containing the communication tools with the user and the HEMS is instantiated (line $2$). It creates its tools' description that we will give to the LLM agent. The algorithm stops when the LLM agent has found and stored every desired parameter which can be checked with the $is\_complete$ function (line $3$). At each step, the LLM agent is instantiated with a prompt template and a toolkit (line $5$). Then, we ask the LLM to solve a task (line $6$). After, the algorithm updates the task list (line $8$). Finally, the HEMS function is called with the retrieved parameters. The HEMS function solves (Eq. \ref{eq:1}) to obtain the user's optimal energy consumption schedule and setpoints for the controllable devices.

\begin{algorithm}[ht!]
\caption{Parameters retrieval algorithm} \label{alg:pipeline}
\begin{algorithmic}[1]
\STATE $stored\_parameters \gets None$
\STATE $toolkit \gets init\_tools(user, stored\_parameters)$
\WHILE{\NOT $is\_done()$}
    \STATE $current\_task \gets task\_list[0]$
    \STATE $agent \gets React(LLM, toolkit)$
    \STATE $output \gets agent(current\_task, n\_iter)$
    \IF {$params\_stored(output)$}
\STATE $task\_list.pop(0)$
\STATE $stored\_parameters.append(output)$
    \ENDIF
\ENDWHILE
\RETURN {$HEMS(stored\_parameters)$}
\end{algorithmic}
\end{algorithm}

\subsection{User} \label{userappendix}

We used the embedding model all-MiniLM-L6-v2, a powerful derivative of the MiniLM developed by Microsoft Research \cite{wang2020minilmdeepselfattentiondistillation} to compute embedded the different textual data (user's answers and the perfect answers). We used Mistral AI instruct model V2 to generate the answers of the LLM user \cite{wolf-etal-2020-transformers}.

All LLM starts by embedding a sentence into a vector before generating the following words of the prompt. In this vector space, two sentences with similar meaning will have a similar vector. To quantify user precision, we want to compute a vector similarity metric between a perfect answer that is concise and respects the wanted format and our user answer. We computed the cosine similarity metric to quantify user answer precision in different difficulty modes \cite{Toshevska_2020}. For example, to the question "How many electric vehicles do you own ?", we define the perfect answer as "I own 2.". For some parameters such as the number of electric vehicles or EVs (see Tab. \ref{tab:params}), a very concise answer might not fully represent the context of the sentence (electric vehicles in the example and not solar panels). Thus, we need to add the context of the answer -which is the question here- to obtain the full meaning of the user's answer. Therefore, to compute user precision, we compute the cosine similarity between the embedding of both the question and the user's answer in all three difficulty modes compared to the embedding of the same question and the perfect answer.

The similarity equation between LLM-user and defined perfect answers is presented in Eq. \ref{eqn:2}. $A(Q)$ is the answer of the user to question $Q$, $A^{*}(Q)$ is the best answer to question $Q$ and $f$ is the embedding function.

\begin{equation}
	cosine(Q) = \frac{f(Q + A(Q)) \cdot f(Q + A^{*}(Q))}{\lVert f(Q+A(Q))\rVert \lVert f(Q+A^{*}(Q)\rVert}
	\label{eqn:2}
\end{equation}

We ask the LLM 20 times eight simple questions -that refer to each one of the eight parameters described in table \ref{tab:params}- in each difficulty level, to get a sample of answers from which we will evaluate the average user precision. For example, to evaluate user precision for parameter $EV$ in difficulty mode "easy E", we give the LLM Fig. \ref{fig:userprompt} (see section \ref{appendixuser} of the appendix) and we ask the LLM this question "How many electric vehicles do you own ?" several times. Then we compute the cosine similarity between the embedding of the question and the generated answers and the embedding of the question and the perfect answer ("I own 2 electric vehicles." in the example). Tab. \ref{tab:precision} shows that the hierarchy between the three difficulty modes is respected. Indeed the answers generated by the LLM user in easy mode are more similar to the optimal answer than those in medium mode.

\begin{figure}[ht]
	\includegraphics[width=0.44\textwidth,height=0.175\textheight]{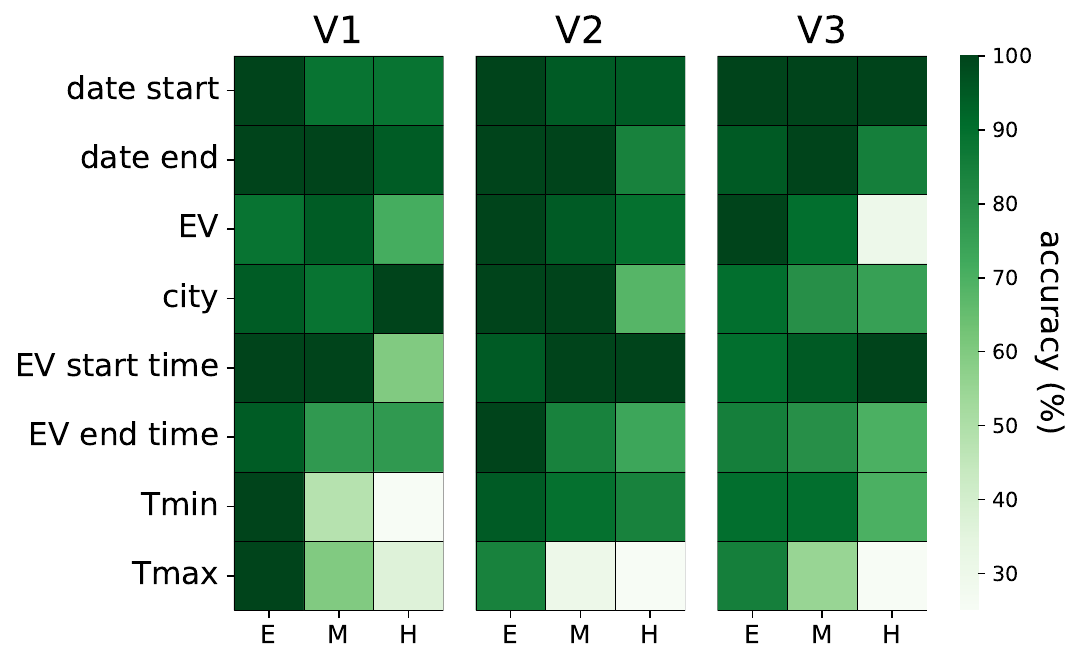}
	\caption[centerfirst]{Average cosine similarity score of user precision \par \small \normalfont This figure shows the cosine similarity score between a typical user answer and a perfect precise answer for each difficulty mode.}
	\label{fig:cossim}
    \Description{This figure is described in the caption.}
\end{figure}

Fig. \ref{fig:cossim} highlights that for all parameters the cosine similarity score of the LLM user difficulty mode Easy (E) is greater than the score in mode Medium (M) except for date parameters. Overall, there is a greater similarity between the user's answers in easy mode and the defined best answers rather than with mode medium and hard. The hierarchy is respected. In the user prompts for easy mode, we ask the LLM user to use the YYYY/MM/DD date format, and in medium mode, we ask the LLM user to use the DD-MM-YYYY format. The date tasks tell the LLM agent to store the date in YYYY/MM/DD format. So the answers in easy mode will not require conversion. Thus, the medium mode requires more work from the LLM agent. However, the embedding model does not take into account those different date formats in its final answer embedding. That is why for date parameters, mode easy and medium scores are close.

\subsection{Results} \label{resultsappendix}

To model the electricity consumption of several appliances of the HEMS, we used the dataset provided in \cite{networkrevolutionHomepageCustomerLed}. We use the parameter settings in Tab. \ref{tab:constant} \cite{6832599} for the temperature evolution formula (Eq. \ref{eq:temp}) with $\eta = 1$ as we use electric heating and not a heat pump and we increased the thermal resistance to model a well-insulated house.

\begin{table}[ht]
  \begin{center}
    \caption{Typical parameter values for residential heating}
    \label{tab:constant}
    \begin{tabular}{c|c|c|c} 
      \textbf{Parameter} & \textbf{Description} & \textbf{Value} & \textbf{Unit}\\
      \hline
      $C_{th}$ & thermal capacitance & 2 & kWh/ \degree C\\
      $R_{th}$ & thermal resistance & 10 & \degree C/kW\\
      $\eta$ & coefficient of performance & 1 & 1\\
    \end{tabular}
  \end{center}
\end{table}

As described in section \ref{methodology} with Fig. \ref{fig:highlevelchart}, the interface starts with step \circled{B1}: the user--LLM agent interaction. To perform this stage, we need to instantiate the LLM agent with its prompt (see appendix Fig. \ref{fig:agentprompt} in section \ref{appendixagent} and section \ref{methodology}) and a task. The LLM agent receives the task "Find the user's city in the United Kingdom and store it (city must be a string)." associated parameter $city$ (see Tab. \ref{tab:params}). The LLM agent will generate Fig. \ref{fig:resultexample}. In line 1, the agent rephrases the main task. In line 2, the agent first decides to fulfill its first goal, i.e. to ask a question to the user in lines 3 to 9. The user answers with line 10. Here the first step \circled{B1} of Fig. \ref{fig:highlevelchart} is completed, the agent has received badly formatted information from the user (instead of just answering "Oxford", the user decided to give extra noisy information). Then, the LLM agent decides to store this information in line 12 and use the tool "store" to do so (lines 13 to 19). It receives a validation message in line 20, completing step \circled{B2} of Fig. \ref{fig:highlevelchart}. Finally, the LLM agent understands that its main task is fulfilled and ends its text generation.

\begin{figure}
    \begin{Verbatim}[fontsize=\footnotesize, breaklines=true, breaksymbol=, numbers=left, xleftmargin=5mm, frame=single]
Task: Find the user's city within the United Kingdom and store it as a string.
Thought: I need to ask the user for his city.
Action:
```
{
"action": "ask_user",
"action_input": "Which city in the United Kingdom do you live in?"
}
```
Observation: I live in Oxford on Banbury Rd.
Thought: I need to store this parameter.
Action:
```
{
"action": "store",
 "action_input": "Oxford"
}
```
Observation: The value was correctly assigned. The task is done
Thought: I have found the user's city and stored it.
Final Answer: The user lives in Oxford.
\end{Verbatim}
\captionof{figure}{LLM agent retrieval example}
\label{fig:resultexample}
\Description{The figure is described in the text.}
\end{figure}

\begin{figure*}[pt]
    \centering
    \begin{subfigure}[b]{0.44\textwidth}
        \centering
        \begin{tikzpicture}
        \node[inner sep=0] (image) {\includegraphics[width=\textwidth, height=0.1\textheight]{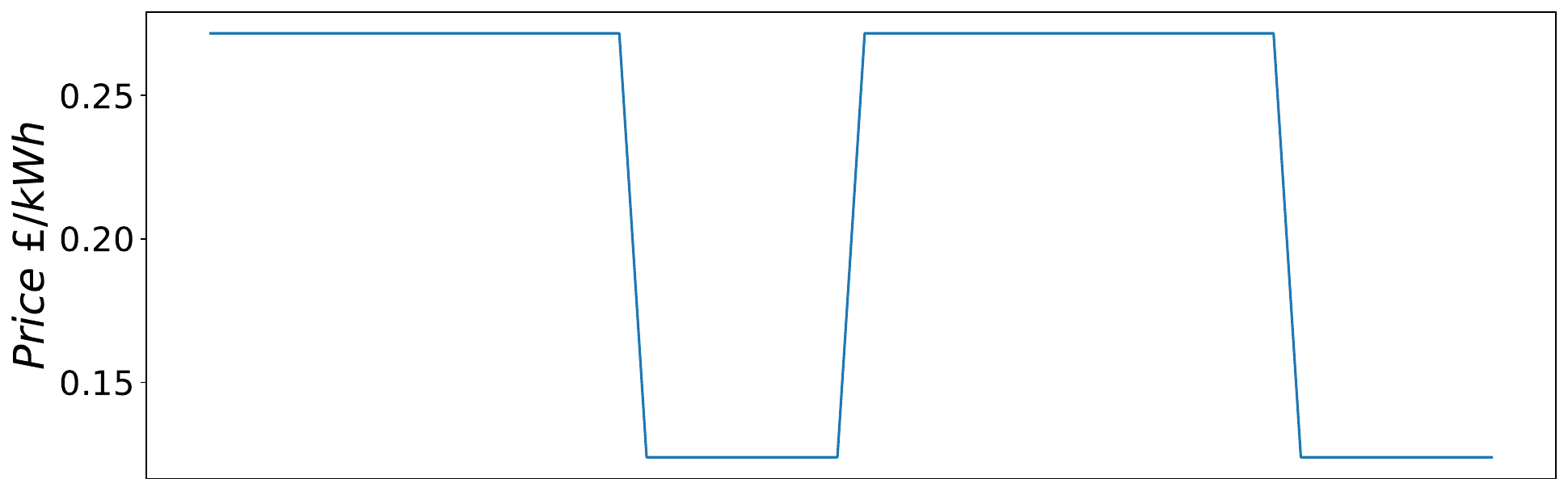}};

        \node[rectangle, draw=red!100,, text width=10.2cm, text height=2.8cm, ultra thick, rounded corners=10] (a) at (2,6) {\begin{minipage}{.85\linewidth}
            \begin{lstlisting}[breaklines=false]
(*{\begin{center} \large \bfseries Agent-User interaction\end{center}}*)
(*\bfseries Agent*) Q1: Where do you live ?
(*\bfseries $\rightarrow$ User*) A1: I live in London on 44 Station Road.
(*\bfseries Agent*) Q2: When do you want the simulation to start ?
(*\bfseries $\rightarrow$ User*) A2: I want the simulation to start on the 16th of September.
(*\bfseries Agent*) Q3: When do you want the simulation to end ?
(*\bfseries $\rightarrow$ User*) A3: I want the simulation to end on the 22th of September.
(*\bfseries Agent*) Q4: How many electric vehicles do you own ?
(*\bfseries $\rightarrow$ User*) A4: I own 2 volvo XC40 and one diesel pickup truck.
(*\bfseries Agent*) Q5: When do you come back from work ?
(*\bfseries $\rightarrow$ User*) A5: I return at 7 PM (UK time) after picking up my kids.
(*\bfseries Agent*) Q6: When do you leave your house ?
(*\bfseries $\rightarrow$ User*) A6: I leave my house at 9 AM (UK time) after a good breakfast.
(*\bfseries Agent*) Q7: What is your house minimum comfort temperature ?
(*\bfseries $\rightarrow$ User*) A7: My house minimum comfort temperature is 18 degrees celsius.
(*\bfseries Agent*) Q8: What is your house maximum comfort temperature ?
(*\bfseries $\rightarrow$ User*) A8: I set my house thermostat at the maximum of 20 degrees.
            \end{lstlisting}
        \end{minipage}};
        \node[rectangle, draw, text width=5cm, text height=1.8cm, ultra thick, rounded corners=10] (b) at (11,7) {\begin{minipage}{.85\linewidth}
            \begin{lstlisting}[breaklines=false]
(*{\begin{center} \large \bfseries Formatted HEMS inputs\end{center}}*)
{
start_date: 16/09/2024,
end_date: 22/09/2024,
EV: 2,
city: London,
EV_start_time: 19:00,
EV_end_time: 9:00,
Tmin: 18,
Tmax: 20,
}
            \end{lstlisting}
        \end{minipage}};
        \node[rectangle, text centered, draw=blue!100, text width=2cm, text height=0.5cm, text depth = 0.15cm,ultra thick, rounded corners=5] (b) at (11,3.5) {\large \bfseries HEMS};
        \draw[->] (7.3, 7) -- (8.3, 7);
        \draw[->] (11, 4.8) -- (11, 4);
        \draw[->] (11, 3) -- (11, 2) -- (4.3, 2) -- (4.3, 1.1);
        \end{tikzpicture}
        \caption{Electric price evolution over time}    
        \label{fig:optim1}
    \end{subfigure}
    \hfill
    \begin{subfigure}[b]{0.54\textwidth}  
        \centering 
        \includegraphics[width=\textwidth, height=0.1\textheight]{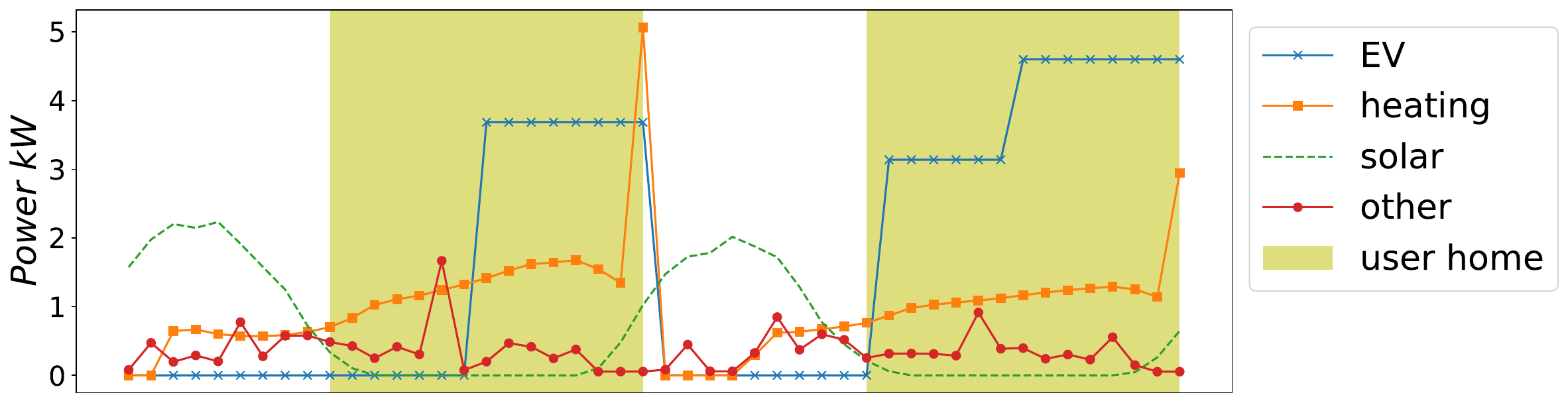}
        \caption{Consumed power evolution over time}    
        \label{fig:optim2}
    \end{subfigure}
    \vskip\baselineskip
    \begin{subfigure}[b]{0.45\textwidth}   
        \centering 
        \includegraphics[width=\textwidth, height=0.1\textheight]{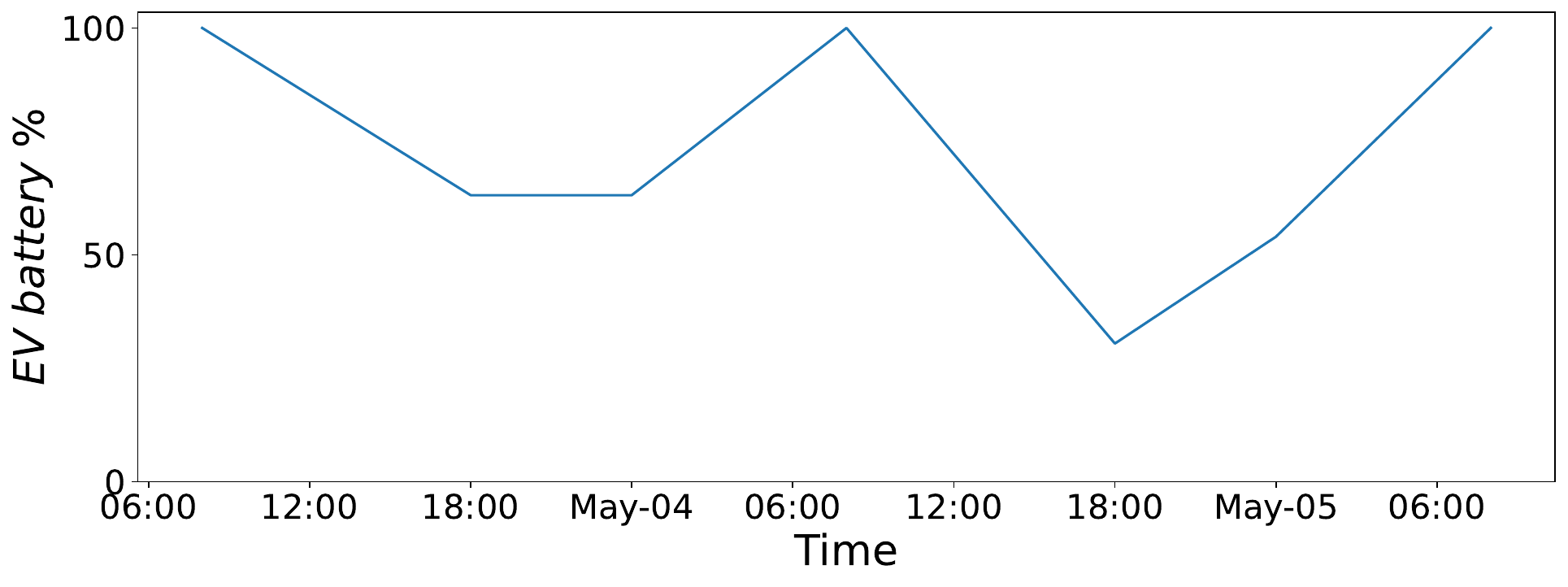}
        \caption{Electric vehicle(s) batteries over time}    
        \label{fig:optim3}
    \end{subfigure}
    \hfill
    \begin{subfigure}[b]{0.54\textwidth}   
        \centering 
        \includegraphics[width=\textwidth, height=0.1\textheight]{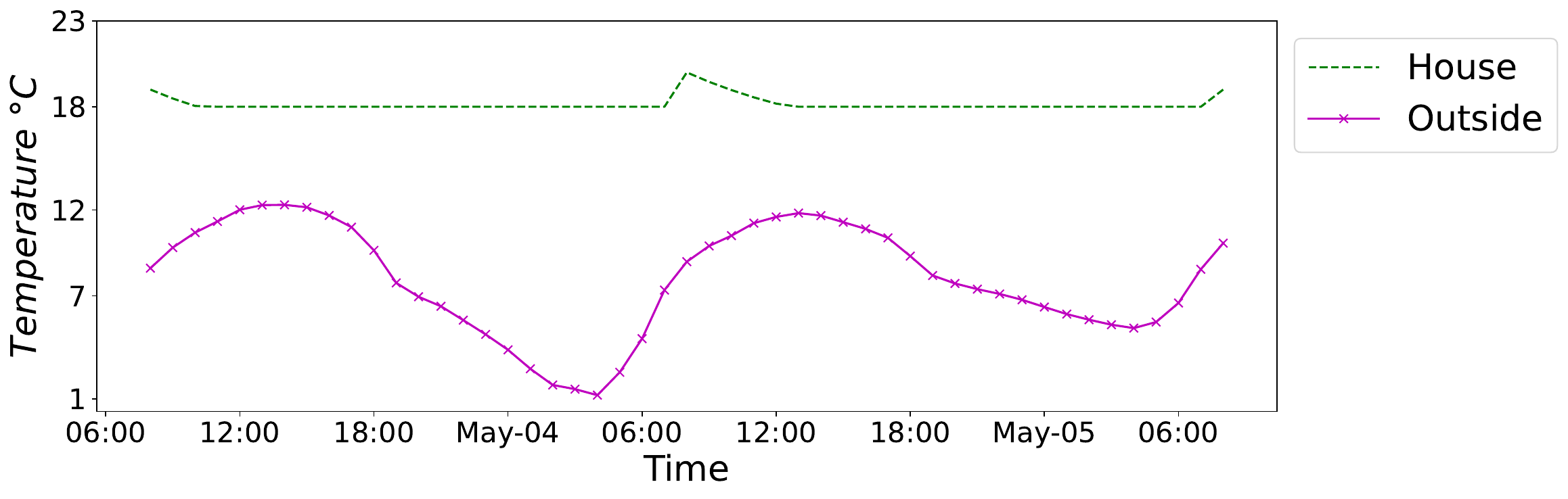}
        \caption[]%
        {Temperature evolution over time}    
        \label{fig:optim4}
    \end{subfigure}
    \caption[centerfirst]{The interaction process of the LLM-integrated HEMS from a user's perspective \par \small \normalfont The \textbf{Agent-User interaction} text box describes a typical usage of the interface. From the user's answers, the LLM agent stores the formatted parameters in text box \textbf{Formatted HEMS inputs}. The \textbf{HEMS} computes the optimal energy consumption schedule in Fig. \ref{fig:HEMS}. Fig. \ref{fig:optim1} depicts the electric price evolution over time which is an Economy 7 setup. Fig. \ref{fig:optim2} describes the consumed power evolution over time for heating (square marker orange), EV charging (x marker blue), other appliances (o marker red), and solar-generated power (dashed green). The yellow shaded areas represent when the user is home or not. Fig. \ref{fig:optim3} represents electric vehicle(s) batteries over time. Fig. \ref{fig:optim4} describes the outside temperature (x marker purple) and house temperature (dashed green) evolution over time.}
    \Description{The figure is fully described in the caption.}
    \label{fig:HEMS}
\end{figure*}

After having successfully retrieved all user information and formatted it in a HEMS-compatible format (see right part of Fig. \ref{HEMS}), the HEMS will compute the optimal consumption energy schedule. Fig. \ref{fig:HEMS} shows the whole user-LLM interaction process from the user's perspective and the HEMS output: the most cost-efficient control of electric appliances while fulfilling the user's constraints. Fig. \ref{fig:optim3} shows that when the user leaves his house the car(s) battery is full. The car is charged when the user is back home. The HEMS charges the car only when the electricity price is low allowing the user to use his car as he wants while diminishing his bills. With Fig. \ref{fig:optim4}, we can check that the temperature constraints are also validated as $T_{house}$ is between $T_{min}$ and $T_{max}$ values. Also, in the low price period, the HEMS heats the house to the $T_{max}$ limit, to heat less when electricity is more expensive.

\begin{figure}[ht]
	\includegraphics[width=0.47\textwidth,height=0.23\textheight]{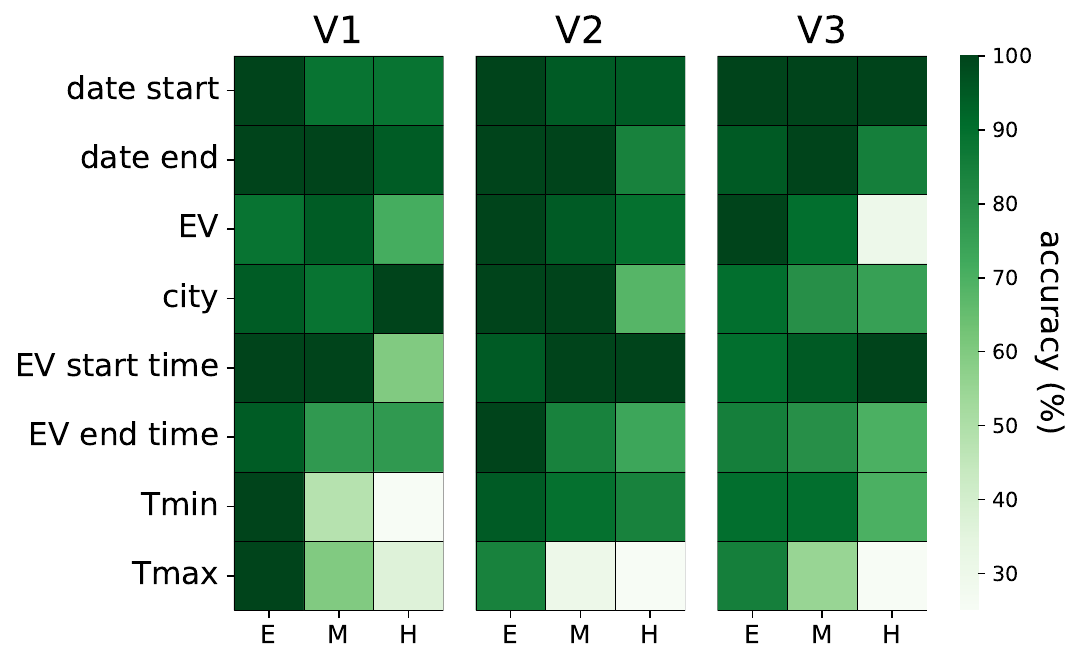}
	\caption[centerfirst]{Accuracy for each parameter with $ReAct+example$\par \small \normalfont This figure represents the rate of retrieval per parameter per difficulty level for each LLM. E, M, and H respectively stand for Easy, Medium, and Hard difficulty levels. The models tend to underperform with temperature parameters ($Tmin$, $Tmax$) and with dates ($date\_end$, $date\_start$).}
    \Description{This figure is described in the caption.}
	\label{fig:accuracyparam}
\end{figure}

Fig. \ref{fig:accuracyparam} show the average accuracy of the $ReAct+example$ agent for each parameter. It can be seen that the LLM agent struggled the most with $Tmin$ and $Tmax$ parameters. The models had trouble handling Celsius degrees and hallucinated by doing conversions even though it was not required in the given tasks. Temperature parameters are described to the LLM user in this way:

\begin{itemize}
    \item easy: minimum house comfort temperature: \$TMIN °C
maximum house comfort temperature: \$TMAX °C
    \item medium: Your house comfort temperature is between \$TMIN °C and \$TMAX °C.
    \item hard: You like to set your house thermostat to be between \$TMIN and \$TMAX degrees Celsius.
\end{itemize}

By mixing the temperature information in one sentence for difficulty mode medium and hard, the LLM user tends to add those two pieces of information or mix them in its answers to the LLM agent questions. That is where the LLM agent fails to extract the precise value. For example to the LLM agent question: "What is your preferred maximum house temperature?", the LLM user in medium or hard mode can answer "My preferred maximum house temperature is between 18 and 19 degrees Celsius". In this particular case, the LLM agent may get confused and store 18.5 instead of 19. Here, the user's answer is imprecise, so the LLM agent interprets that any values between 18 and 19 are correct. However, the LLM agent should understand that it needs to ask a second question to clarify the user's preference. From the different tests we have run, we observed that the LLM user tends to be more concise when asked about $Tmin$ which greatly helps the LLM agent. We conjecture that the LLM user is more precise with $Tmin$, because the minimum temperature is the main concern when heating your house, which may be more expressed in the datasets used by LLM developers to train LLMs.

\begin{figure*}[ptbh]
\centering
\resizebox{\textwidth}{!}{\begin{tabular}{@{}ccc@{}}
\parbox[b]{.1\textwidth}{\textcolor{white}{V}} Number of questions &  
\textcolor{white}{V12} Duration (s) &  \textcolor{white}{2} Number of iterations\\
\raisebox{55pt}{\parbox[b]{.05\textwidth}{V1}}
\subfloat[]{\includegraphics[width=0.3\textwidth]{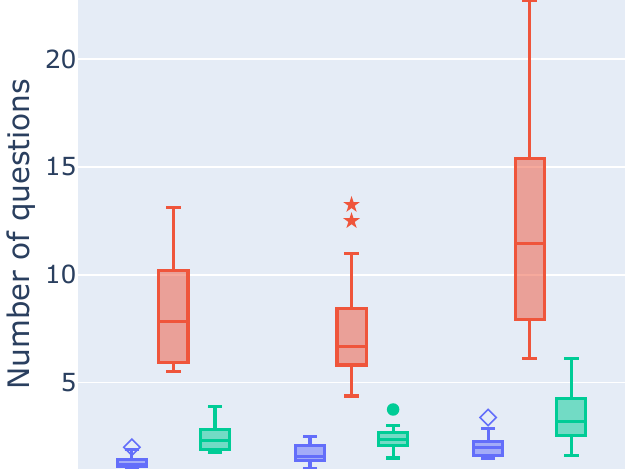}\label{fig:a}} & 
\subfloat[\label{fig:b}]{\includegraphics[width=0.3\textwidth]{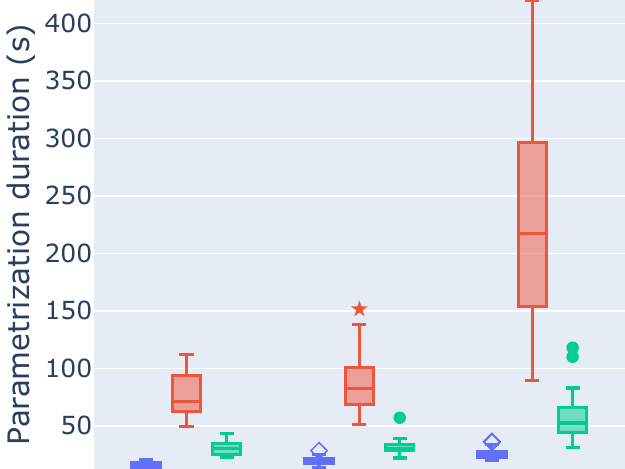}} &
\subfloat[\label{fig:c}]{\includegraphics[width=0.3\textwidth]{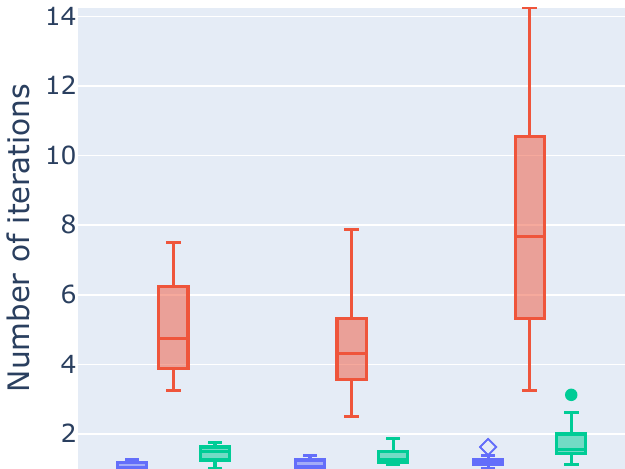}} \\
\raisebox{55pt}{\parbox[b]{.05\textwidth}{V2}} \subfloat[\label{fig:d}]{\includegraphics[width=0.3\textwidth]{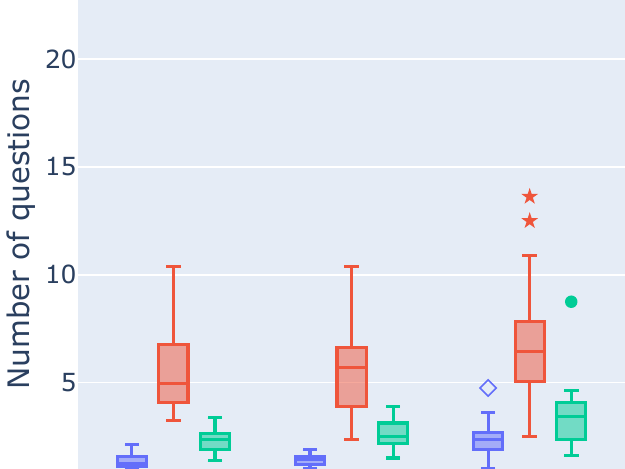}} &
\subfloat[\label{fig:e}]{\includegraphics[width=0.3\textwidth]{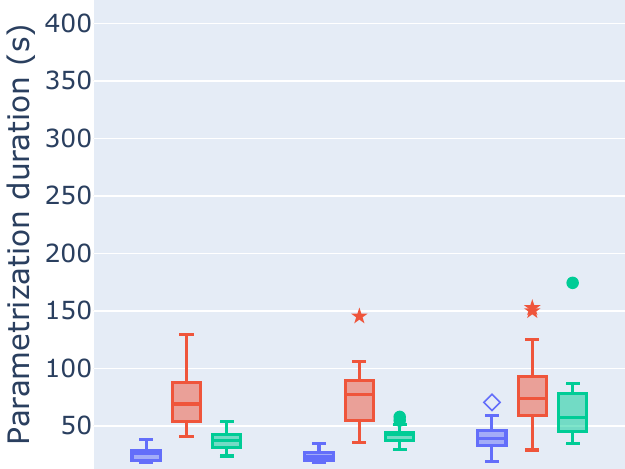}} &
\subfloat[\label{fig:f}]{\includegraphics[width=0.3\textwidth]{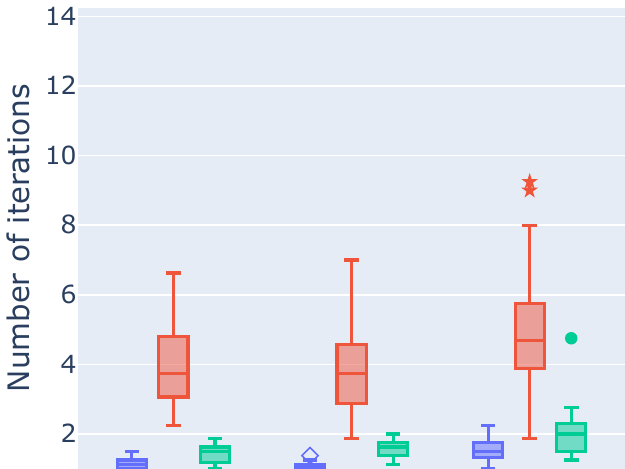}}\\
\raisebox{55pt}{\parbox[b]{.05\textwidth}{V3}} \subfloat[\label{fig:g}]{\includegraphics[width=0.3\textwidth]{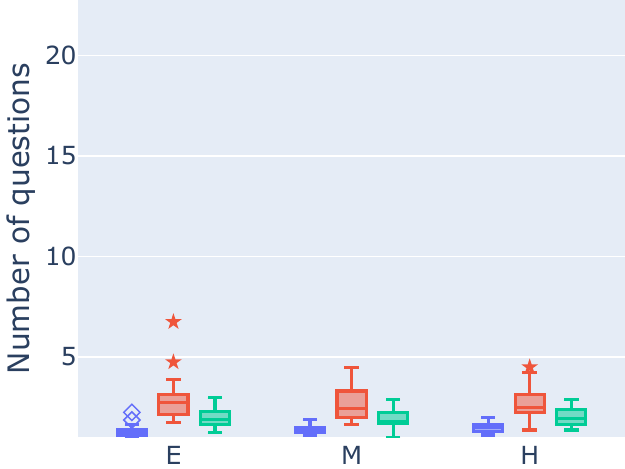}} &
\subfloat[\label{fig:h}]{\includegraphics[width=0.3\textwidth]{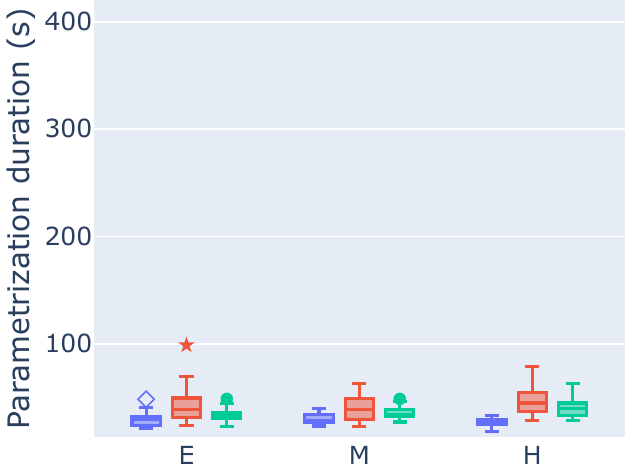}} &
\subfloat[\label{fig:i}]{\includegraphics[width=0.3\textwidth]{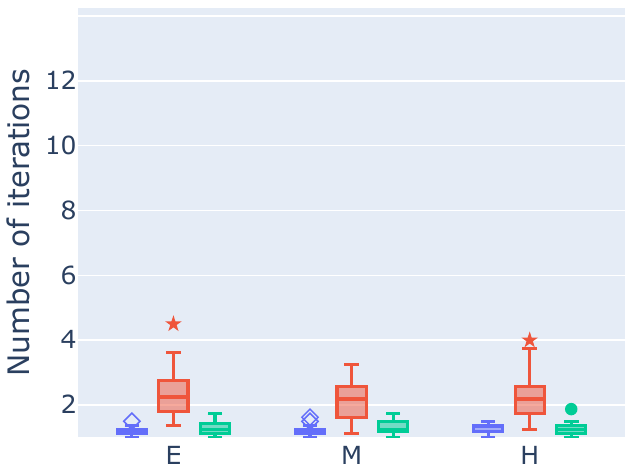}}\\
\\
\multicolumn{3}{c}{\parbox[b]{.05\textwidth}{\textcolor{white}{V12}} \includegraphics[width=0.4\textwidth]{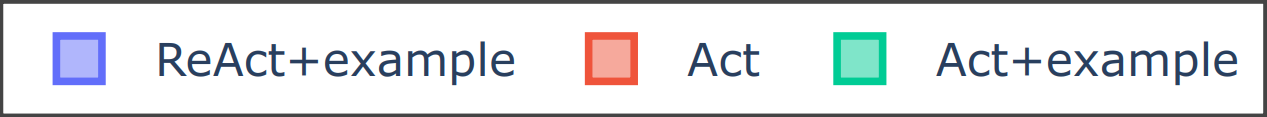}}\\
\end{tabular}}
        
\caption[centerfirst]{Boxplot distribution of test results for model V2 \par \small \normalfont Subgraphs of column 1 (Fig. \ref{fig:a}, \ref{fig:d}, \ref{fig:g}) represent the boxplot distribution of the number of questions asked by the model to retrieve a \textbf{single} parameter for each difficulty level with three different agent types. Subgraphs of column 2 (Fig. \ref{fig:b}, \ref{fig:e}, \ref{fig:h}) depict the boxplot distribution of a \textbf{single} parameter retrieval duration in seconds per difficulty level. Subgraphs of column 3 (Fig. \ref{fig:c}, \ref{fig:f}, \ref{fig:i}) represent the boxplot distribution of the number of algorithms iterations required by the agent to retrieve a \textbf{single} parameter for each difficulty level. Subgraphs of row k (1,2 or 3) refers to the results obtained with LLM version Vk. Agent types $ReAct+example$, $Act$, and $Act+example$ are symbolized by blue diamonds, red stars, and green circles.}
\Description{Figure described in the caption.}
\label{fig:results}
\end{figure*}

\section{Prompts} \label{prompts}

\subsection{LLL-agent prompts} \label{appendixagent}

In this section, we describe the agent prompt template of Fig. \ref{fig:agentprompt}, used for the $ReAct+example$ agent type. As mentioned in section \ref{react}, we first describe the main goal of the LLM agent (line 1). We chose to keep this first line simple for a shorter prompt. Then, the prompt continues with the tools list and their description (lines 1 and 5). Here the value \$TOOL\_DESCRIPTION is replaced by Fig. \ref{fig:tool_description}. The format rules it must follow for function calling are described on lines 3 to 13. We added two redundant messages on the type correctness of the "action\_input" key after having experienced several failures due to this error. The LLM had trouble handling double quotes when working with strings. Line 5 was added to make the LLM remember the importance of using correct action names in the JSON blob. The ReAct rules on lines 15 to 27, follow a classic ReAct implementation. We used the keywords "Observation:", and "Final Answer:" as generation-stopping criteria, respectively to parse the LLM-generated text and then call the appropriate function, and to check if the task is successful.

\begin{figure}[ht!]
\begin{Verbatim}[fontsize=\small, frame=single, breaklines=true, breaksymbol=, numbers=left, xleftmargin=5mm]
{'action': 'ask_user', 'action_description': "Return the user's answer to a query", 'action_input': 'query: A question to a user', 'action_output': "The user's answer"}
{'action': 'store', 'action_description': 'Store a value by assigning it to an argument', 'action_input': 'value: The value to store.', 'action_output': 'A validation message'}
\end{Verbatim}
\captionof{figure}{Tool description}
\label{fig:tool_description}
\Description{The figure is described in the text.}
\end{figure}

Fig. \ref{fig:agentprompt} is then formatted into a chat template segmenting the agent prompt template, the example, and the current task. Such a format allows the LLM to understand its goals, examples, and current task fully. Fig. \ref{fig:chat_template} describes the used chat template. Each part starts with the keyword "<|im\_start|>" and ends with the keyword "<|im\_end|>" to make the LLM understand the chat template segmentation. Each part is then composed of a role (system, agent, or assistant) defining the purpose of the part. "System" refers to LLM agent instructions. "Agent" is the LLM agent, so each text generated by the LLM agent will have the role "agent". "Assistant" is the role giving tasks. It is not an LLM but a part of the initialized prompt. For example, for each HEMS parameter retrieval, the "Assistant" could be: "Now you need to retrieve the minimum temperature". After the role comes the content: \$PROMPT\_TEMPLATE, \$TASK\_EXAMPLE, \$EXAMPLE, \$TASK, respectively replaced by the actual initialized agent prompt template of Fig. \ref{fig:agentprompt}, a task example like "Find the user's number of solar panels and store it (number must be an integer).", how should the LLM solve this task, the current task the LLM should solve. For agent types $Act$ and $Act+example$ the agent prompt template is shorter (no need to describe the thinking process at each step) and the chat template is without example for agent type $Act$.

\begin{figure}[ht!]
    \begin{Verbatim}[fontsize=\small, frame=single, breaklines=true, breaksymbol=, numbers=left, xleftmargin=5mm]
<|im_start|>system
$PROMPT_TEMPLATE<|im_end|>
<|im_start|>agent
$TASK_EXAMPLE<|im_end|>
<|im_start|>assistant
$EXAMPLE<|im_end|>
<|im_start|>agent
$TASK<|im_end|>
<|im_start|>assistant
    \end{Verbatim}
    \captionof{figure}{Agent chat template}
    \label{fig:chat_template}
    \Description{The figure is described in the text.}
\end{figure}

\begin{figure}[ht!]
\begin{Verbatim}[fontsize=\small, frame=single, breaklines=true, breaksymbol=, numbers=left, xleftmargin=5mm]
You will have to solve tasks as best you can. To do so, you have access to the following tools: $TOOLS_DESCRIPTION

The way you use the tools is by specifying a json blob. Specifically, this json should have a `action` key (with the name of the tool to use) and a `action_input` key (with the input to the tool going here). The `action_input` key must be a valid python object (string, integer, list, float). If your `action_input` is a string you must use "" instead of ''.

The only values $TOOL_NAME that should be in the "action" field are: $TOOLS_LIST

The $JSON_BLOB should only contain a SINGLE action and MUST be formatted as markdown, do NOT return a list of multiple actions. Here is an example of a valid $JSON_BLOB:
{
  "action": $TOOL_NAME,
  "action_input": $INPUT
}

Make sure to have the $INPUT in the right format for the tool you are using, and is a correct JSON input. The $JSON_BLOB must be formatted as markdown and only use a SINGLE action at a time.

ALWAYS use the following format:
Task: The input task you must answer.
Thought: You should always think about one action to solve the task.
Action: $JSON_BLOB
Observation: The result of the action.

This Thought/Action/Observation can repeat N times, you  should take several steps when needed. You should keep repeating the above format until you have enough information to answer the question without using any more tools.

You must always end your output with the following format:
Thought: I now know the final answer.
Final Answer: The final answer to the original input question.

Now begin! Reminder to ALWAYS use the exact characters `Final Answer: $ANSWER` when you provide a definitive answer.
\end{Verbatim}
\captionof{figure}{Agent prompt template}
\label{fig:agentprompt}
\Description{The figure is described in the text.}
\end{figure}

In case the JSON blob parsed from the text generated by the LLM agent is not correct, the parser will return an error message (see Fig. \ref{fig:error_message}). This message sums up the LLM agent's mistake, replacing keyword \$ERROR with the actual Python error message (line 
2). Then two recalls are given on type and function name correctness.

\begin{figure}[ht!]
\begin{Verbatim}[fontsize=\small, frame=single, breaklines=true, breaksymbol=, numbers=left, xleftmargin=5mm]
You made a mistake in your JSON blob.
Here is the Python error message: $ERROR

Remember, `action_input` must be a valid Python object (str, int, list, float).
Try again with the right format and the right keys.
\end{Verbatim}
\captionof{figure}{Error message}
\label{fig:error_message}
\Description{The figure is described in the text.}
\end{figure}

\subsection{LLL-user prompts} \label{appendixuser}

In Fig. \ref{fig:userprompt}, \ref{fig:userpromptmedium} and \ref{fig:userprompthard}, \$CITY, \$EV, \$TMIN, \$TMAX, \$ARRIVAL\_TIME, \$LEAVING\_TIME, \$DATE1 and \$DATE2 are replaced with the random values for each test in a format that depends on difficulty level (for example, with date values, "2024/10/18" in easy mode and "October, 18th, 2024" in hard mode). All three prompts start with the main task of the LLM user (line 1). Then, we describe the user personal information we want our LLM user to answer with (Fig. \ref{fig:userprompt} lines 4 to 11, Fig. \ref{fig:userpromptmedium} lines 4 to 10, and Fig. \ref{fig:userprompt} lines 4 to 9). Along those lines, we give the same information to each LLM user in all three difficulty modes, but we change the format of this information and we have added noisy information in harder modes so that the user answers will contain too much information for the agent and its parameter retrieval will be harder. Finally, we add each prompt with rules to follow when answering (Fig. \ref{fig:userprompt} lines 13 to 14, Fig. \ref{fig:userpromptmedium} lines 12 to 13, and Fig. \ref{fig:userprompt} lines 11 to 12). Notice how the rule precision diminishes with the rising difficulty. We even force the LLM user to generate a more complex answer in Fig. \ref{fig:userpromptmedium} and \ref{fig:userprompthard}.

\begin{figure}[ht!]
\begin{Verbatim}[fontsize=\small, frame=single, breaklines=true, breaksymbol=, numbers=left, xleftmargin=5mm]
You will have to answer questions about your personal information.
Here is your personal information:

city: $CITY
number of electric vehicles: $EV
minimum house comfort temperature: $TMIN C
maximum house comfort temperature: $TMAX C
time when you come back home: $ARRIVAL_TIME
time when you leave your house: $LEAVING_TIME
simulation start date: $DATE1
simulation end date: $DATE2

Answer with a concise first-person sentence of less than 25 tokens.
For date questions, use this format YYYY/MM/DD.
\end{Verbatim}
\captionof{figure}{User prompt template easy mode}
\label{fig:userprompt}
\Description{The figure is described in the text.}
\end{figure}

\begin{figure}[ht!]
\begin{Verbatim}[fontsize=\small, frame=single, breaklines=true, breaksymbol=, numbers=left, xleftmargin=5mm]
You will have to answer questions about your personal information.
Here is your personal information:

You live in $CITY  in England, in a house.
You own $EV volvo XC40.
Your house comfort temperature is between $TMIN °C and $TMAX °C.
You come back from work at $ARRIVAL_TIME PM (UK time) after picking up your kids.
You leave your house at $LEAVING_TIME AM (UK time) after a good breakfast.
You want the simulation to start on the $DATE1.
You want the simulation to end on the $DATE2.

Answer with a concise first-person sentence.
For date questions, use this format dd-mm-yyyy.
\end{Verbatim}
\captionof{figure}{User prompt template medium mode}
\label{fig:userpromptmedium}
\Description{The figure is described in the text.}
\end{figure}

\begin{figure}[ht!]
\begin{Verbatim}[fontsize=\small, frame=single, breaklines=true, breaksymbol=, numbers=left, xleftmargin=5mm]
You will have to answer questions about your personal information.
Here is your personal information:

You live in $CITY in England on Banbury Road, in a house with your family.
You own $EV volvo XC40, one diesel pickup truck and a gas-powered  motorcycle.
You like to set your house thermostat to be between $TMIN and $TMAX degrees Celsius.
You are back from work at $ARRIVAL_TIME PM due to traffic.
You go to work at $LEAVING_TIME AM to escape traffic.
You want to simulate your electric consumption between $DATE1 and $DATE2.

You MUST Answer with two first-person sentences for each question.
Add an explanation to your answer.
\end{Verbatim}
\captionof{figure}{User prompt template hard mode}
\label{fig:userprompthard}
\Description{The figure is described in the text.}
\end{figure}

Like the LLM agent, we use a chat template for the LLM user (see Fig. \ref{fig:chat_template_user}). Here, such a format allows the LLM to fully understand its goals, and the question it must answer. We follow the same format: start keyword, role (system, agent, and user), content, and end keyword. \$USER\_PROMPT, and \$QUERY are respectively changed by one of the user prompts (Fig. \ref{fig:userprompt}, \ref{fig:userpromptmedium}, and \ref{fig:userprompthard}).

\begin{figure}[ht!]
\begin{Verbatim}[fontsize=\small, frame=single, breaklines=true, breaksymbol=, numbers=left, xleftmargin=5mm]
<|im_start|>system
$USER_PROMPT<|im_end|>
<|im_start|>agent
$QUERY<|im_end|>
<|im_start|>user
\end{Verbatim}
\captionof{figure}{User chat template}
\label{fig:chat_template_user}
\Description{The figure is described in the text.}
\end{figure}

\section{Online Resources} \label{online}
The study's codes are available on a GitHub repository \cite{LLM}. The README.md file of this repository describes the project's architecture, installation, and usage.

\end{document}